%% file: cmr.tex
\documentclass[10pt,twocolumn,letterpaper]{article}

\usepackage[pagenumbers]{cvpr} % To force page numbers, e.g. for an arXiv version

\input{preamble}

\newcommand{\bfsection}[1]{\noindent\textbf{#1.}}
\definecolor{cvprblue}{rgb}{0.21,0.49,0.74}
\usepackage[pagebackref,breaklinks,colorlinks,citecolor=cvprblue]{hyperref}
\usepackage{algorithm}
\usepackage{algorithmic}
\usepackage{multirow}
\usepackage{booktabs}
\usepackage{graphicx}
\usepackage{colortbl}
\usepackage{bm}
\def\paperID{6881} % *** Enter the Paper ID here
\def\confName{CVPR}
\def\confYear{2024}

\title{Split to Merge: Unifying Separated Modalities for \\ Unsupervised Domain Adaptation}

\author{Xinyao Li$^1$ \quad Yuke Li$^2$\footnotemark[1] \quad  Zhekai Du$^1$ \quad Fengling Li$^3$ \quad Ke Lu$^1$ \quad Jingjing Li$^{1}$\footnotemark[1]\\
$^1$University of Electronic Science and Technology of China\\
$^2$Boston College\\
$^3$University of Technology Sydney\\
{\tt\small xinyao326@outlook.com, lidwh@bc.edu, zhekaid@std.uestc.edu.cn} \\
{\tt\small fenglingli2023@gmail.com, kel@uestc.edu.cn, lijin117@yeah.net}
}
\begin{document}
\maketitle
%\renewcommandf{\thefootnote}{\fnsymbol{footnote}}
\footnotetext[1]{Corresponding author.}
\input{sec/0_abstract}    
\input{sec/1_intro}
\input{sec/2_related}

\input{sec/3_method_bk}

\input{sec/4_experiments}

\input{sec/X_suppl}

{
    \small
    \bibliographystyle{ieeenat_fullname}
    \bibliography{main}
}


\end{document}

%% file: preamble.tex
\usepackage[dvipsnames]{xcolor}

%% file: sec/0_abstract.tex
\begin{abstract}
    %Large vision-language models~(VLMs) have shown impressive  downstream generalization and zero-shot capabilities. However, existing VLM transfer methods either focus on fine-tuning the language branch or prompting the visual branch, often neglecting the intrinsic relationships and distinctions between the two modalities. 
    %To fully harness the rich multimodal knowledge in large pretrained VLMs like CLIP, in this paper we present Cross-Domain Modality Separation~(CDMoS) for unsupervised domain adaptation. Inspired by recent advances in modality gap, we design a lightweight modality separation network to explicitly disentangle the CLIP-extracted features into language-associated and vision-associated components. To promote the exchange of modality-invariant information while preserving modality-specific details, we propose a Modality-Ensemble Training (MET) algorithm, which cooperatively updates both sets of feature components. Finally, we achieve adaptation without target supervision by aligning cross-domain features from both modalities via a modality discriminator. We thoroughly analyze and validate our method on three challenging benchmarks. Results demonstrate that our method is concrete and efficient, achieving new SOTA with surprisingly low computational overheads.
    
    Large vision-language models (VLMs) like CLIP have demonstrated good zero-shot learning performance in the unsupervised domain adaptation task. Yet, most transfer approaches for VLMs focus on either the language or visual branches, overlooking the nuanced interplay between both modalities. In this work, we introduce a Unified Modality Separation (UniMoS) framework for unsupervised domain adaptation. Leveraging insights from modality gap studies, we craft a nimble modality separation network that distinctly disentangles CLIP's features into language-associated and vision-associated components. Our proposed Modality-Ensemble Training (MET) method fosters the exchange of modality-agnostic information while maintaining modality-specific nuances. We align features across domains using a modality discriminator. Comprehensive evaluations on three benchmarks reveal our approach sets a new state-of-the-art with minimal computational costs. Code: \href{https://github.com/TL-UESTC/UniMoS}{https://github.com/TL-UESTC/UniMoS}.
    
    \end{abstract}

%% file: sec/1_intro.tex
\section{Introduction}
\label{sec:intro}

Unsupervised domain adaptation (UDA) \cite{ben2010theory,cdan,dann,rtn} aims to apply knowledge trained on a source domain to an unlabeled target domain, a process invaluable in data-scarce scenarios. Conventional methods often struggle with bridging the gap between source and target domains, finding it challenging to develop consistent features across domains \cite{aaa,mcd}. In image classification, aligning vision features while neglecting semantic content can lead to difficulties in differentiating complex samples \cite{dapl,adclip}. 
Vision-language models (VLMs) such as CLIP \cite{clip} and ALIGN \cite{align} circumvent these issues through joint multimodal pretraining on images and texts. This extensive pretraining endows publicly available VLMs with robust zero-shot transfer abilities and a broad base of conceptual knowledge, making them highly suitable for comprehensive UDA. They facilitate alignment across both visual and textual modalities, enhancing adaptability and applicability in diverse contexts.

\begin{figure}[t]
  \centering
  \includegraphics[width=0.47\textwidth]{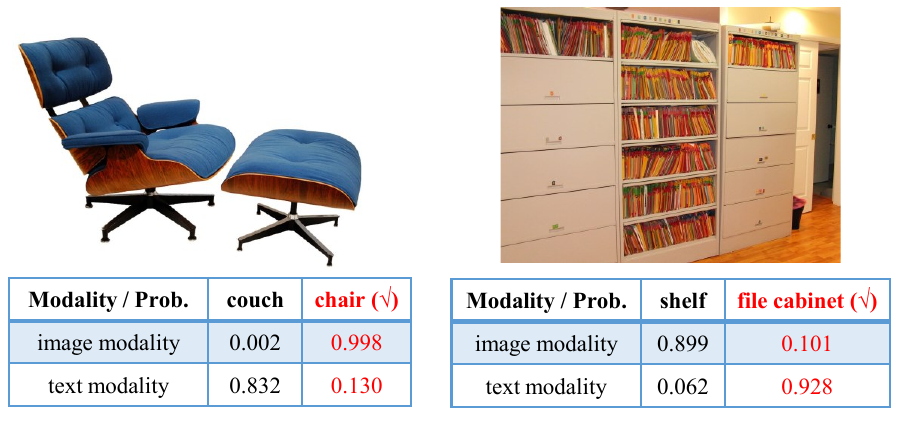}
  \vspace{-6pt}
  \caption{Examples of modality-specific information from task Art$\rightarrow $RealWorld in Office-Home dataset. The digits are top-2 highest classification probabilities given by both modalities. }
  \vspace{-12pt}
  \label{fig1}
\end{figure}

Previous studies have shown promising results by adapting VLMs like CLIP for unsupervised domain adaptation (UDA). For instance, DAPrompt \cite{dapl} introduces learning both domain-agnostic and domain-specific text embeddings, while PADCLIP \cite{padclip} focuses on fine-tuning the vision branch of CLIP for adaptive visual feature extraction. However, recent research \cite{liang2022mind,jiang2023understanding} highlights a \textit{modality gap} in VLMs, revealing that, despite training efforts, vision and text features often remain distinctly distributed. We argue that adapting a single modality is less than ideal due to the existence of unique, modality-specific cues in misaligned textual and visual components. We suggest that certain samples are best classified using specific modalities, a hypothesis supported by empirical observations in \cref{fig1}. This figure shows differing classification patterns when each modality is adapted independently to the unlabeled target data. Text modality results derive from CLIP’s zero-shot capabilities, while image results come from linear probing with target pseudo-labels. For instance, visually straightforward items like a cushioned chair are accurately classified by the vision linear classifier after tuning on target dataset. However, pretrained CLIP can erroneously categorize such items under visually similar classes. In contrast, complex items with nuanced semantic details, like a file cabinet resembling a shelf, may confuse the vision classifier, while CLIP’s broader knowledge base facilitates correct zero-shot predictions. In summary, while the vision branch effectively discerns class-specific visual patterns, the text branch leverages semantic information to clarify ambiguous cases. This observation lays the groundwork for a multimodal adaptation framework that synergistically combines the strengths of both modalities.
%For an easy example of a chair, the vision classifier gives highly confident correct prediction while CLIP tends to classify it as a couch. A file cabinet that looks like a shelf easily fools the vision classifier but is correctly identified by CLIP. We deduce the reason is that the multimodal pretraining of CLIP gives it rich semantic information of different classes, which allows it to perform accurate classification on samples with complex semantic information (i.e., a shelf-like cabinet containing files). 
%But without tuning on target dateset, it may misclassify simple samples due to lack of dataset-specific vision information. Instead, a trained linear classifier exposed to target data can capture inter-class vision relationships unique to a given dataset, improving discrimination between similar classes like chairs and couches. Therefore, we ask a natural question: \textit{Can we take the best of both modalities for UDA?}

A direct approach to domain adaptation involves concurrently fine-tuning vision branch and crafting textual prompts, which risks disturbing the image-text representation pairs in pretrained CLIP and is computationally intensive \cite{clip-adapter, tip-adapter}. As a more efficient alternative, we propose to explicitly disentangle CLIP-extracted visual features into two complementary parts. The first component retains the language-associated semantic knowledge inherent in CLIP, while the second focuses on vision-specific attributes crucial for distinguishing between nuanced visual categories.

We  devised a set of modality separation networks with dual branches to project CLIP-encoded visual features into distinct language-associated components (LAC) and vision-associated components (VAC). An orthogonal regularization  is employed to ensure these branches yield discrete, disentangled representations. Each component is optimized based on its inherent modality strengths. For the LAC branch, we utilize knowledge distillation on target data to harness the rich semantic content from the original pretrained CLIP model. Additionally, we implement a debiasing method to mitigate dataset bias in CLIP's zero-shot results. For the VAC branch, the locality structure within visual feature spaces \cite{nrc,elpt,nrc++} is leveraged to generate visual pseudo-labels for supervised learning on target data. We then introduce a novel Modality-Ensemble Training (MET) strategy that synergistically merges outputs from both modalities. A weight generator dynamically assembles these predictions, supervised by VAC pseudo-labels on target data and actual labels on source data. Importantly, the text modality output remains isolated during MET to preserve independent training and maintain pretrained semantics. Additionally, a modality discriminator is utilized to align LAC and VAC across domains for unsupervised domain adaptation. Trained on source data to distinguish between LAC and VAC, this discriminator is frozen on the target domain, directly updating the separation networks to produce domain-invariant LAC and VAC. This approach ensures a consistent modality separation across domains, facilitating simultaneous adaptation in both modalities.

\textbf{Contributions:} 1. We investigate the modality gap phenomenon in the context of applying Vision-Language Models (VLMs) to unsupervised domain adaptation, revealing the limitations of adapting a single modality;
2. We introduce a novel framework, Unified Modality Separation (UniMoS), which, coupled with a Modality-Ensemble Training (MET) approach, facilitates effective multimodal adaptation;
3. Our comprehensive analysis and validations underscore  and efficiency of the proposed UniMoS, demonstrating its ability to set new state-of-the-art benchmarks while maintaining low computational demands.

%% file: sec/2_related.tex
\section{Related work}
\label{sec:related}
\textbf{Unsupervised domain adaptation (UDA).} A core challenge in UDA is aligning representations between the source domain and unlabeled target domain.  Prior  techniques can be categorized as discrepancy-based~\cite{dan,mdd,mdd-long} and adversarial methods~\cite{dann,cdan,mcd}. Discrepancy-based methods explicitly minimizes divergence metrics including MMD~\cite{mdd}, MDD~\cite{mdd-long}, etc. Adversarial methods extract domain invariant features via a min-max game between the feature extractor and domain discriminator~\cite{dann,cdan}. Recent works focus on exploiting  target data structures via self-training techniques~\cite{icon,can,nrc,nrc++,eidco}. ICON~\cite{icon} learns an invariant classifier with consistent predictions to remove the spurious correlation inconsistency in the target domain. EIDCo~\cite{eidco} combines Mixup~\cite{mixup} with IDCo loss~\cite{chen2020simple,he2020momentum} to explore target data distribution. Vision transformer (ViT)~\cite{vit} and its variants have also gained popularity due to their superior performance~\cite{pmtrans,cdtrans,tvt}. PMTrans~\cite{pmtrans} mixes patch representations in SwinTransformer~\cite{swin} as an intermediate domain bridge. CDTrans~\cite{cdtrans} aligns features extracted by DeiT~\cite{deit} via cross-attention.

\begin{figure*}[t]
  \centering
  \includegraphics[width=0.95\textwidth]{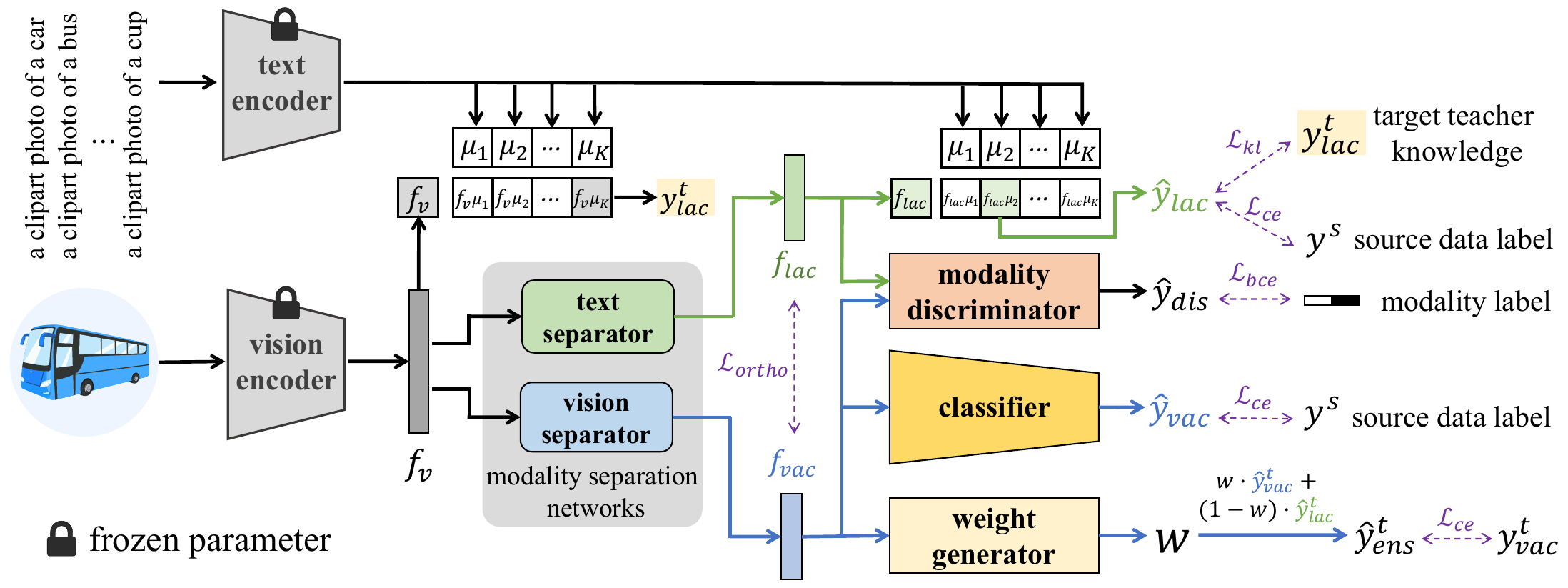}
  \vspace{-8pt}
  \caption{Framework of our method. We freeze the pretrained vision and text encoder of CLIP. CLIP-extracted vision features are  disentangled into language-associated components ($f_{lac}$) and vision-associated components ($f_{vac}$) by the modality separation networks. We obtain zero-shot results from CLIP as teacher knowledge, and distill the knowledge to LAC. We then introduce a weight generator to assemble the modality outputs to train VAC. A  modality discriminator is applied to align LAC and VAC from both domains. }  % 
  %\caption{Framework of our method. CLIP-extracted vision features are first disentangled into language-associated components (LAC) and vision-associated components (VAC). We design different paradigms to train LAC and VAC, and introduce a weight generator to assemble the final output. A cross-domain modality discriminator is applied to align LAC and VAC from both domains. Best viewed in color.}
  \label{fig2}
  \vspace{-8pt}
\end{figure*}

\textbf{Vision-language models} (VLMs) have shown great generalization abilities due to extensive multimodal pretraining~\cite{clip,align,wang2022ofa,wang2021ufo}. CLIP~\cite{clip} is trained from 400 million text-image pairs, while ALIGN~\cite{align} leverages more than one billion text-image pairs. Subsequent works have built on pretrained VLMs in various ways. Some learn prompt texts to transfer VLMs to downstream tasks~\cite{coop,cocoop,dapl,paiss2023teaching,denseclip}, while others incorporate additional tunable layers on the frozen pretrained encoder~\cite{clip-adapter,tip-adapter}. Beyond utilizing existing VLMs, research also aims to improve VLM training~\cite{jiang2023understanding,slip}. Liang \textit{et al.}~\cite{liang2022mind} reveal that VLMs exhibit a modality gap, failing to perfectly align multimodal features. Jiang \textit{et al.}~\cite{jiang2023understanding} conduct theoretical analysis on modality gap and propose latent space regularization to preserve modality-specific information. MaPLe~\cite{maple} utilize prompt learning on both modality branches to improve alignment. Our approach is fundamentally different  since we disentangle VLM-extracted features posteriorly instead of training VLM from scratch, requiring far less computation costs. Besides, our method requires no labeling on target domain. VLMs have also been adopted for UDA~\cite{dapl,adclip,padclip}. DAPrompt~\cite{dapl} proposes to learn domain-specific and domain-agnostic textual prompts for each class. AD-CLIP~\cite{adclip}  learns domain invariant prompts by conditioning on image style and content features. PADCLIP~\cite{padclip} dynamically adjusts learning rate while tuning the CLIP vision branch to prevent catastrophic forgetting. However, these methods perform adaptation on either the visual or textual modality in isolation. Our work aims to address this limitation by proposing a unified adaptation framework of the multimodal features.

%% file: sec/3_method_bk.tex
\section{Method}

\subsection{Problem formulation}
\label{formulation}
In this study, superscripts differentiate domains, with symbols lacking superscripts applicable to both domains. We consider a labeled source domain $\mathcal{D}^s=\{(x_i^s,y_i^s)\}_{i=1}^{N^s}$ and aim to develop a model generalizable to an unlabeled target domain $\mathcal{D}^t=\{(x_i^t)\}_{i=1}^{N^t}$. CLIP~\cite{clip} features a vision encoder $g_{vis}$ and a text encoder $g_{txt}$. The vision feature for an image input $x$ is denoted as $f_v=g_{vis}(x)$. Employing the zero-shot inference strategy from~\cite{padclip}, we construct naive prompts $\{(t_i)\}_{i=1}^K$ as \textit{a [DOMAIN] photo of a [CLASS]}, with $K$ representing the number of classes, \textit{[DOMAIN]} indicating domain specifics, and \textit{[CLASS]} the class name. Text features are then derived as $\mu_i=g_{txt}(t_i)$. $\mu_i$ and $f_v$ are both features with $d_v$ dimension. Classification is based on the highest cosine similarity between $f_v$ and $\mu_i$: 
\begin{equation} 
    \hat{y}_{zs}=\underset{i}{\arg\max} \; \cos(\mu_i, f_v). 
    \label{zeroshot}
\end{equation}

\cref{zeroshot} may not be ideal for unlabeled target data due to the existence of modality gap \cite{liang2022mind}. 
To tackle this, we conceptualize vision inputs as a composite of a vision-associated component (VAC) and a language-associated component (LAC), denoted as $f_v = \{f_{vac}, f_{lac}\}$. This leads us to obtain modality-specific classification results $y_{vac}$ and $y_{lac}$, before constructing a cross-modality output:
\begin{equation}
y_{ens} = w \cdot y_{vac} + (1-w) \cdot y_{lac},
\label{ensemble1}
\end{equation}
where $w$ is a set of learnable weights harmonizing the contributions of VAC and LAC. This design seeks to balance modality-agnostic information sharing and modality-specific information capturing.

Instead of separating LAC and VAC during training, we utilize $y_{ens}$ to guide VAC learning, incorporating complementary modality information for a holistic cross-modal training. Furthermore, a fixed weight $w$ may lack flexibility across diverse datasets and scenarios, potentially obscuring the distinction between VAC and LAC.
Addressing this, we introduce a dynamic $w$ that adeptly discriminates between modalities within $y_{ens}$, calibrating their influence in the training. This strategy ensures tailored training approaches for different datasets or training stages, facilitating modality-specific information utilization. Next we detail on the training and aligning of LAC and VAC.

\subsection{Modality separation networks}

We first introduce the modality separation networks that disentangles CLIP-extracted features, which comprise two separators as shown in \cref{fig2}. These networks partition CLIP-extracted vision features into LAC and VAC using the text separator $G_{txt}$ and the vision separator $G_{vis}$, respectively. The separated components are defined as $f_{lac}=G_{txt}(f_v)$ and $f_{vac}=G_{vis}(f_v)$. Both separators are linear layers preserving the dimensionality of $f_v$, such that $f_{vac}, f_{lac} \in \mathbb{R}^{d_v}$. Drawing on deep feature separation principles \cite{bousmalis2016domain}, we apply an orthogonal loss to maintain the distinctness of LAC and VAC:
\begin{equation}
\mathcal{L}_{ortho} = |f_{lac}^{s} \cdot {f_{vac}^{s}}^{\top } |_{F}^{2} + |f_{lac}^{t} \cdot {f_{vac}^{t}}^{\top } |_{F}^{2}.
\label{ortho}
\end{equation}

Different outputs for LAC and VAC are then generated. For the text modality, we utilize the zero-shot inference  of CLIP to classify LAC by calculating the cosine similarity between LAC and CLIP's text features, forming logits:
\begin{equation}
\hat{y}_{lac} = (\hat{l}_1, \hat{l}_2, \cdots \hat{l}_k), \quad \hat{l}_i = \cos(\mu_{i}, f_{lac})/T,
\label{lac_hat}
\end{equation}
where $T$ is temperature in pretrained CLIP.
For VAC, we route it through a linear classifier with layers $\Phi_{1} \in \mathbb{R}^{d_v \times d_b}$ and $\Phi_2 \in \mathbb{R}^{d_b \times K}$, producing the bottleneck feature with dimension $d_b$ and output via:
\begin{equation}
f_b = \Phi_1 (f_{vac}), \quad \hat{y}_{vac} = \Phi_2 (f_b).
\label{vac_hat}
\end{equation}
We provide implementation details  in Supplementary.

\subsection{Modality-ensemble training}
Having obtained  disentangled components, we design customized training paradigm for each modality. A learnable weight further connects both modalities, establishing  a unified modality-ensemble training framework.

\bfsection{Learning  LAC} 
To preserve the rich semantic content in pretrained CLIP, we distill this knowledge to LAC.
For the target data, zero-shot similarity scores derived from pretrained CLIP serve as the teacher knowledge:
\begin{equation}
y_{lac}^t = (l_1-\overline{l} , l_2-\overline{l}, \cdots, l_k-\overline{l}), \quad l_i = \cos(\mu_{i}, f_{v}^t)/T,
\label{lac}
\end{equation}
with $\overline{l}= \frac{1}{K} \sum_{k=1}^{K} l_k$ normalizing the CLIP outputs and $T$ the temperature of pretrained CLIP. The teacher knowledge in \cref{lac} guides the distillation for the unlabeled target LAC, while for the source data, cross-entropy loss is applied directly using labeled source data. The overall training loss for LAC combines \cref{lac_hat} and \cref{lac} as follows:
\begin{equation}
\mathcal{L}_{lac} = \mathrm{KL}(\hat{y}_{lac}^t,  y_{lac}^t) + \alpha \mathrm{CE}(\hat{y}_{lac}^s,  y^s),
\label{llac}
\end{equation}
where $\alpha$ adjusts the influence of source data supervision, $\mathrm{KL}(\cdot,\cdot)$ is the Kullback-Leibler divergence, and $\mathrm{CE}(\cdot,\cdot)$ represents the standard cross-entropy loss.

\bfsection{Obtaining pseudo label for VAC} 

Focusing on image modality, we aim to enhance the locality structure of vision representations—high inter-class discriminability and tight intra-class distribution—a feature that CLIP-extracted vision features lack, as detailed in \cref{tsne_init}. To instill these locality structures within VAC, we utilize a K-means-based deep clustering approach \cite{deepcluster,shot} to generate pseudo-labels for unlabeled target data. We calculate the clustering centroids for class $k$ as follows:
\begin{equation}
\phi_k = \frac{\sum_{x^t} \delta_k (\mathrm{softmax} (\hat{y}_{ens}^t)) \cdot f_b^t}{\sum_{x^t} \delta_k (\mathrm{softmax} (\hat{y}_{ens}^t))},
\end{equation}
where $\hat{y}_{ens}^t$ represents target ensemble outputs discussed below, and $\delta_k$ selects the $k_{th}$ logit. To mitigate imbalances in text modality predictions from CLIP~\cite{padclip, debiaspl}, we implement Approximated Controlled Direct Effect (ACDE)~\cite{debiaspl} to adjust similarity scores obtained in \cref{lac_hat}:
\begin{align}
& \tilde{y}_{lac}^t = \hat{y}_{lac}^t - \tau \log \hat{p}, \
& \hat{p} \leftarrow m\hat{p} + (1-m)\frac{1}{B}\sum_{i=1}^{B}p_i,
\label{debias1}
\end{align}
where $m$ is momentum, $\tau$ is a debiasing factor, $B$ is the batch size, and $p_i=\mathrm{softmax}(\hat{y}_{lac}^t)$ denotes classification probability of LAC. The ensemble outputs, used in the centroid calculation, are then defined as $\hat{y}_{ens}^t=w \cdot \hat{y}_{vac}^t + (1-w) \cdot \tilde{y}_{lac}^t$.
For any given target bottleneck feature $f_b^t$, we compute its cosine similarity with all centroids, assigning the class with the highest similarity as the pseudo-label:
\begin{equation}
y_{vac}^t = \arg \underset{k}{\max} \, \cos(f_b^t, \phi_k).
\label{cluster}
\end{equation}

\bfsection{Learning VAC}
We now train vision component on unifies outputs from both modalities.
Utilizing \cref{debias1} and \cref{vac_hat}, the target ensemble output $\hat{y}_{ens}^t$ is formulated as:
\begin{equation}
\hat{y}_{ens}^t = w \cdot \hat{y}_{vac}^t + (1-w) \cdot \tilde{y}_{lac}^t,
\label{ensemble2}
\end{equation}
with the weight $w=W(VAC^t)$ produced by the weight generator $W$, as depicted in \cref{fig2}. Referring to \cref{formulation}, we optimize $\hat{y}_{ens}^t$ rather than $\hat{y}_{vac}^t$ directly, with $\tilde{y}_{lac}^t$ serving as an auxiliary in training VAC and thus detached from the computational graph in \cref{ensemble2}.

To enhance individual discriminability and global diversity, thereby preserving the locality structure of vision representations, we follow state-of-the-art~\cite{shot,ahmed2021unsupervised,elpt} to apply an information maximization loss $\mathcal{L}_{im}$ comprising two components. The entropy loss $\mathcal{L}_{ent}$ improves individual certainty:
\begin{equation}
\mathcal{L}_{ent} = -\mathbb{E}_{x^t \in \mathcal{D}^t}\left[\sum_{k=1}^{K} \delta_k (\hat{y}_{ens}^t) \log \delta_k (\hat{y}_{ens}^t)\right],
\end{equation}
and the diversity loss fosters diverse class distributions:
\begin{equation}
\mathcal{L}_{div} = -\sum_{k=1}^{K} \overline{q}_k \log \overline{q}_k,
\end{equation}
where $\overline{q}_k=-\mathbb{E}_{x^t \in \mathcal{D}^t} \delta_k(\hat{y}_{ens}^t)$. Hence, $\mathcal{L}_{im}$ is defined as:
\begin{equation}
\mathcal{L}_{im} = \mathcal{L}_{ent} - \mathcal{L}_{div}.
\label{im}
\end{equation}

The training of VAC  is supervised by target pseudo labels for the vision modality, obtained through \cref{cluster}, while source labels directly optimize $\hat{y}_{vac}^s$:
\begin{equation}
\mathcal{L}_{vac} = CE(\hat{y}_{ens}^t, y_{vac}^t) + \beta CE(\hat{y}_{vac}^s, y^s) + \mathcal{L}_{im},
\label{lvac}
\end{equation}
where $\beta$ modulates the impact of source data supervision.

\subsection{Aligning source and target by discriminator}
To achieve domain adaptation on both modalities, we introduce a modality discriminator $D$ to align VAC and LAC from both domains. Our approach utilizes a singular modality discriminator trained on the source domain to differentiate LAC from VAC, and then assesses alignment on the target domain. Proper alignment across domains would enable $D$ to discern LAC and VAC on the target domain without direct training. The modality discriminator is trained using binary cross-entropy loss:
\begin{equation}
\mathcal{L}_{bce} = -[y_{dis} \log \hat{y}_{dis} + (1-y_{dis}) \log (1-\hat{y}_{dis})],
\label{dis}
\end{equation}
where $y_{dis}$ represents the modality label (0 for VAC, 1 for LAC) and $\hat{y}_{dis}$ is the output of $D$. 

Notably, $D$ is only trained on the source domain using \cref{dis}. On the target domain, only the separators $G_{vis}$ and $G_{txt}$ are updated to minimize \cref{dis}, aligning target LAC and VAC with the source ones.

\subsection{Training and inference}
\label{training}
\bfsection{Training}
As depicted in \cref{fig2}, the pretrained text encoder and vision encoder are frozen, and we  optimize parameters of $G_{txt}$, $G_{vis}$, $\Phi_1$, $\Phi_2$, $W$, $D$, denoted as $\theta_{G_{txt}}$, $\theta_{G_{vis}}$, $\theta_{\Phi_1}$, $\theta_{\Phi_2}$, $\theta_{W}$, $\theta_{D}$, respectively. 
%Combining Eq. (\ref{llac},\ref{ortho},\ref{lvac},\ref{dis}), 
Combining \cref{llac},  \cref{ortho}, \cref{lvac}, \cref{dis}, we define the following optimization problem:
\begin{align}
    & \theta_{G_{txt}} = \underset{\theta_{G_{txt}}}{\arg \min} \; \mathcal{L}_{lac} + \gamma\mathcal{L}_{ortho} + \gamma\mathcal{L}_{bce}, \label{opt}\\
    & \theta_{G_{vis}} = \underset{\theta_{G_{vis}}}{\arg  \min} \; \mathcal{L}_{vac} + \gamma\mathcal{L}_{ortho} + \gamma\mathcal{L}_{bce}, \nonumber \\
    & \theta_{W}, \, \theta_{\Phi_1}, \, \theta_{\Phi_2} = \underset{\theta_{W},\theta_{\Phi_1},\theta_{\Phi_2}}{\arg \min} \; \mathcal{L}_{vac}, \nonumber \\
    & \theta_{D} = \underset{\theta_{D}}{\arg \min} \; \gamma \mathcal{L}_{bce} \nonumber ,
\end{align}
where $\gamma$ is hyperparameter controlling regularization terms $\mathcal{L}_{bce}$ and $\mathcal{L}_{ortho}$. We present detailed training procedure of UniMoS in Supplementary. 

\bfsection{Inference} 
%During inference, we adopt the $y_{ens}^t$ in \cref{ensemble2} as final prediction result on target data. We adopt fixed weights $w$ during inference. The goal during inference is to obtain higher accuracy, and according to our theory, both modalities contribute to better classification (\cref{fig1}). Therefore we mix outputs from both modalities to take the best from both worlds.
At inference, the final mixed prediction on target data is obtained using $\hat{y}_{ens}^t$ from \cref{ensemble2}, with a fixed mixup weight $w$. The objective is to maximize accuracy by leveraging the strengths of both modalities for improved classification, as supported by our observations (\cref{fig1}), thus integrating outputs from both modalities to harness their combined advantages.

%% file: sec/4_experiments.tex
\newcommand{\hhs}{@{\hspace{11pt}}}
\input{tab/tab_1}

\input{tab/tab_2}

\input{tab/tab_3.tex}

\input{tab/tab_4.tex}

\input{tab/tab_6.tex}

\section{Experiments}
\subsection{Datasets and implementation details}
We extensively evaluate our method on three mainstream UDA benchmarks. \textbf{Office-Home}~\cite{officehome} consists of 65 categories divided into 4 distinct domains. On \textbf{VisDA-2017}~\cite{visda}, the goal is to transfer knowledge from 152k synthetic images (source domain) to 55k images of real items (target domain). \textbf{DomainNet}~\cite{domainnet} is the most challenging UDA benchmark so far, containing 0.6 million samples from 345 categories divided into 6 distinct domains. Following previous works, we additionally provide results on \textbf{Mini-DomainNet}~\cite{saito2019semi,litrico2023guiding,zhang2023rethinking}, a subset of DomainNet with 4 domains and 126 categories.

We conduct all experiments on an NVIDIA RTX 2080Ti GPU. Since our method does not involve updating CLIP's pretrained parameters or prompts, the CLIP-extracted vision and text features are obtained via one single forward and saved in memory, thus greatly saving computation costs. For all tasks, we adopt SGD optimizer with batch size 32, and set  momentum $m$ in \cref{debias1} to 0.99 and debias factor $\tau$ in \cref{debias1} to 0.5. For Office-Home and DomainNet, we train 50 epochs with initial learning rate 3e-3 and adopt annealing strategy~\cite{cosineannealing} for learning rate decay. We train for 10 epochs with initial learning rate 9e-4 on VisDA due to fast convergence. The fixed mixup weight described in \cref{training} is set to 0.3 for all tasks. We set regularization weight $\gamma$ in \cref{opt} to 0.01 across all datasets.

\subsection{Benchmark results}
\label{main_result}
\bfsection{Office-Home} \cref{officehome} gives classification accuracies on 12 adaptation tasks on Office-Home using the ResNet50~\cite{resnet} backbone. To ensure a fair comparison, we categorize CLIP-based methods into two groups: `none-tuning' and \textcolor{gray}{`full-tuning'}. The former involves learning prompts or additional modules without adjusting the pretrained CLIP backbones, while the latter optimizes the pretrained parameters of CLIP for specific tasks. It is evident from the results that our proposed UniMoS consistently outperforms both `none-tuning' and \textcolor{gray}{`full-tuning'} methods.  Notably, we obtain +1.3\% performance boost than the strong baseline PADCLIP, which fine-tunes the CLIP vision backbone. Our method requires no parameter update or data forwarding of CLIP backbones, thus is much computationally  cheaper. Especially on tasks that take P as target domain, we achieve up to +5.4\% performance boost than PADCLIP, demonstrating the superiority of multimodal adaptation. 

\bfsection{VisDA-2017} \cref{visda} shows class-wise classification accuracies on VisDA using ResNet101~\cite{resnet}. Our method achieves the best performance  among `none-tuning' CLIP methods, while slightly falling behind PADCLIP. The reason is that CLIP has not been trained on the synthetic images like those from source domain of VisDA, resulting in incompatibilities between the VisDA dataset and CLIP. Similar observations are made by PADCLIP~\cite{padclip}, which opts to fine-tune the vision branch of CLIP to address this challenge. Nevertheless, our approach outperforms typical UDA methods.

\bfsection{DomainNet} \cref{domainnet} presents classification accuracies of 30 cross-domain adaptation tasks on the most challenging benchmark DomainNet. Rows represent source domains and columns represent target domains. Our method reaches comparable performance with the strong baseline PADCLIP. One significant observation is that our UniMoS obtains lower accuracy than PADCLIP (6.6\% lower in average) on tasks with qdr as target. This discrepancy arises from a significant domain gap between qdr and other domains, which is challenging to bridge without tuning the CLIP backbones. However, despite this, we manage to achieve superior performance to PADCLIP on all other tasks, thereby mitigating the overall 6.6\% performance drop. We also surpass all competing methods significantly. \cref{mini_domainnet} compares UniMoS with available CLIP-based none-tuning methods on Mini-DomainNet, where our method achieves significant performance boost. Detailed results on Mini-DomainNet are given in Supplementary.

\subsection{Ablation study}
In this section we validate the efficacy of each module in UniMoS. \cref{ablation} presents averaged accuracies on Office-Home and VisDA-2017 by  removing specific modules while maintaining other settings identical. The `w/o debiasing' is obtained by skipping the debias procedure in \cref{debias1}. A primary observation is that the removal of any module leads to a performance drop to varying degrees, underscoring the positive contribution of each module to the overall outcome. The `w/o learnable weight' row is obtained by replacing $w$ in \cref{ensemble2} with a constant 0.5, resulting in the most pronounced performance decline. This emphasizes the significance of dynamic weights, which enables VAC to focus on vision-specific parts. Further insights into the effects of dynamic $w$ are detailed in \cref{discussion}. 

\cref{backbones} ablates on the choice of backbones. We experiment with three backbones on Office-Home, and our UniMoS consistently outperforms all competing methods, proving that our method is generalizable across various models. Detailed results are given in Supplementary.
%Modality discriminator has the least impact on final results. We deduce the reason is that the extensive pretraining of CLIP enables it with strong generalization abilities, which mitigates domain gap to some extent.

\subsection{Discussions}
\label{discussion}
\begin{figure}[t]
  \centering
  
  \includegraphics[width=0.38\textwidth]{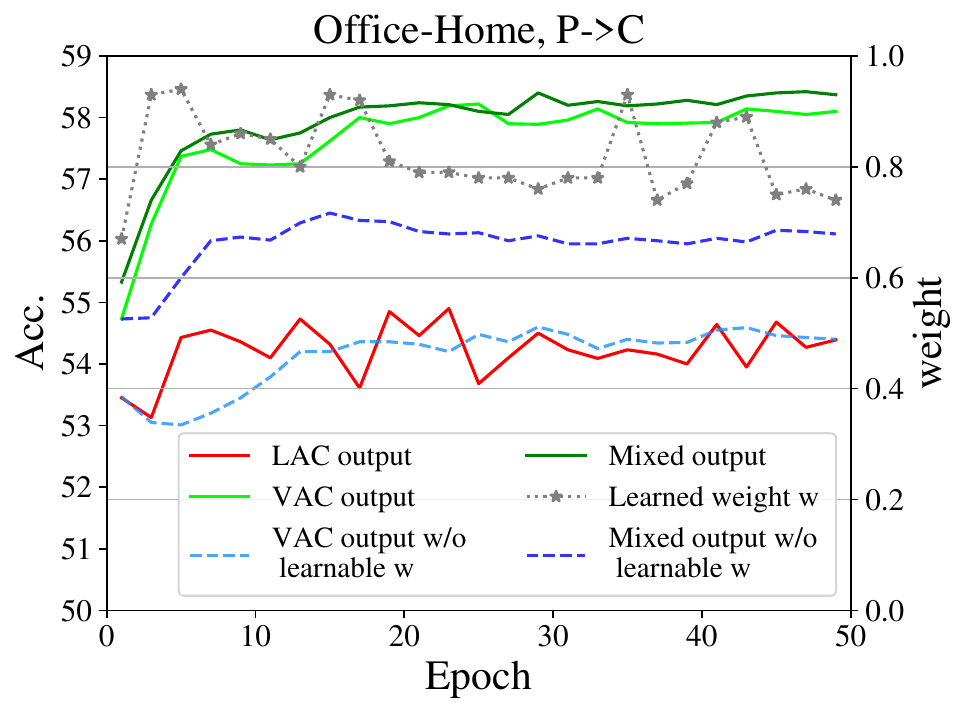}
  %\caption{P$\to$C}
  %\label{conv_1}
 
  \vspace{-8pt}
  \caption{Effects of learnable ensemble weight $w$ on Office-Home.}
  \vspace{-12pt}
  \label{conv}
\end{figure}

\bfsection{Effectiveness of learnable weight $w$} To better understand how the learnable ensemble weight $w$ in \cref{ensemble1} boosts performance, \cref{conv} compares the training process with and without dynamic $w$. We set a fixed weight of 0.5 for both LAC and VAC in `w/o learnable w', and learn dynamic weights (shown as `Learned weight w' in the figure) in other settings. Our first observation is that in our full design, accuracy of VAC increases steadily as the training progresses. Leveraging complementary modality-specific information from LAC, the final mixed outputs achieve higher accuracy than VAC alone. 
As stated in \cref{formulation}, the goal of dynamic ensemble weight is to adaptively identify and preserve modality-specific information. We show in \cref{conv} that, the learned weight changes in each epoch to fit the training process. Without its support (w/o learnable w), the accuracy of VAC outputs drops significantly and finally converges with LAC. This occurs because employing a static weight would compromise the modality separation effects, causing both modalities to collapse to poor performance.  In the example given by \cref{conv}, accuracy of `VAC output w/o learnable w' is 3.5\% lower than that of the full design `VAC output'. The phenomenon indicates the significance of training with dynamic ensemble weights in our design. More examples are given in Supplementary.

\begin{figure*}[t]
  \centering
  \begin{subfigure}{0.32\textwidth}
      \includegraphics[width=\textwidth]{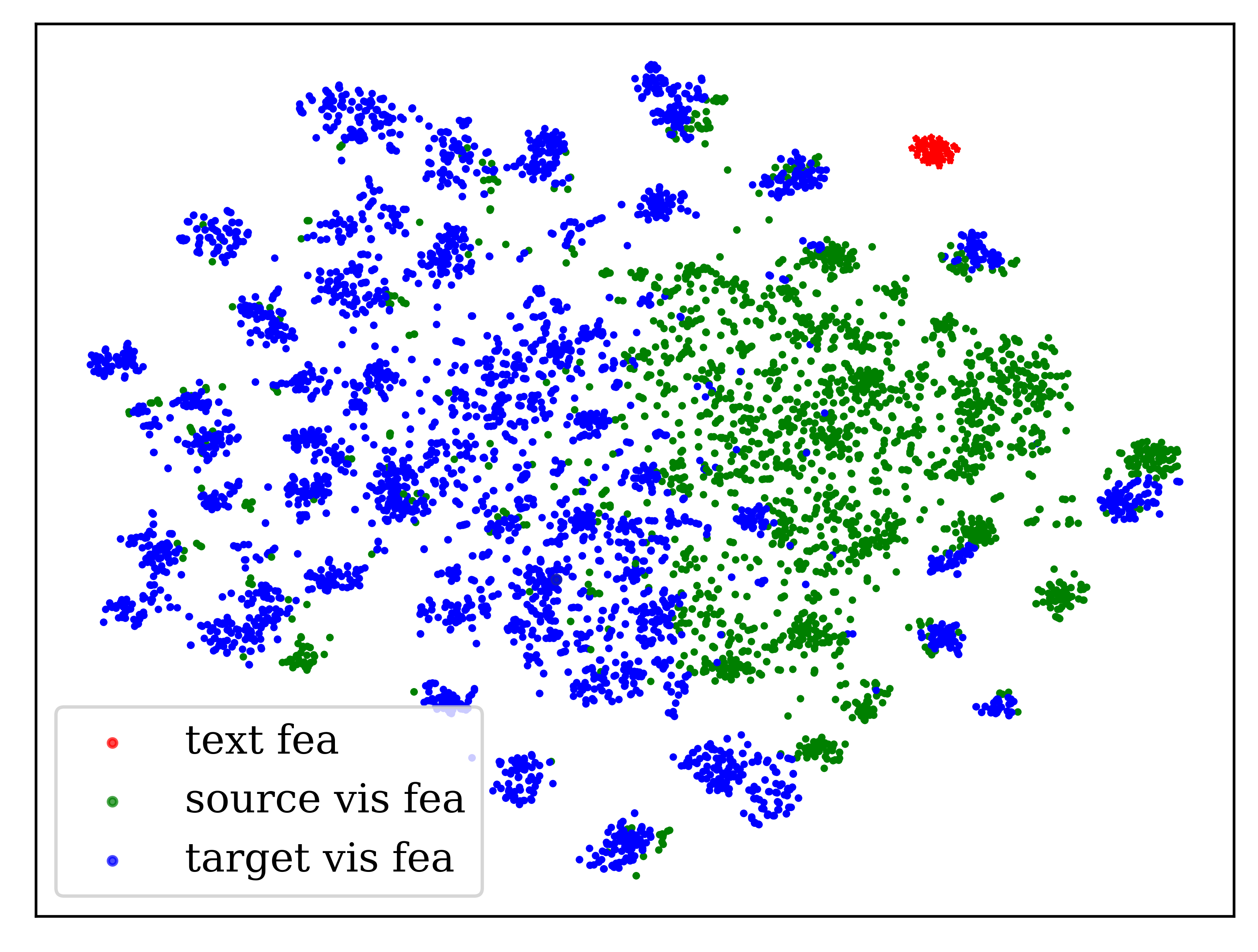}
      \caption{CLIP-extracted  feature distribution.}
      \label{tsne_init}
  \end{subfigure}
  \begin{subfigure}{0.32\textwidth}
      \includegraphics[width=\textwidth]{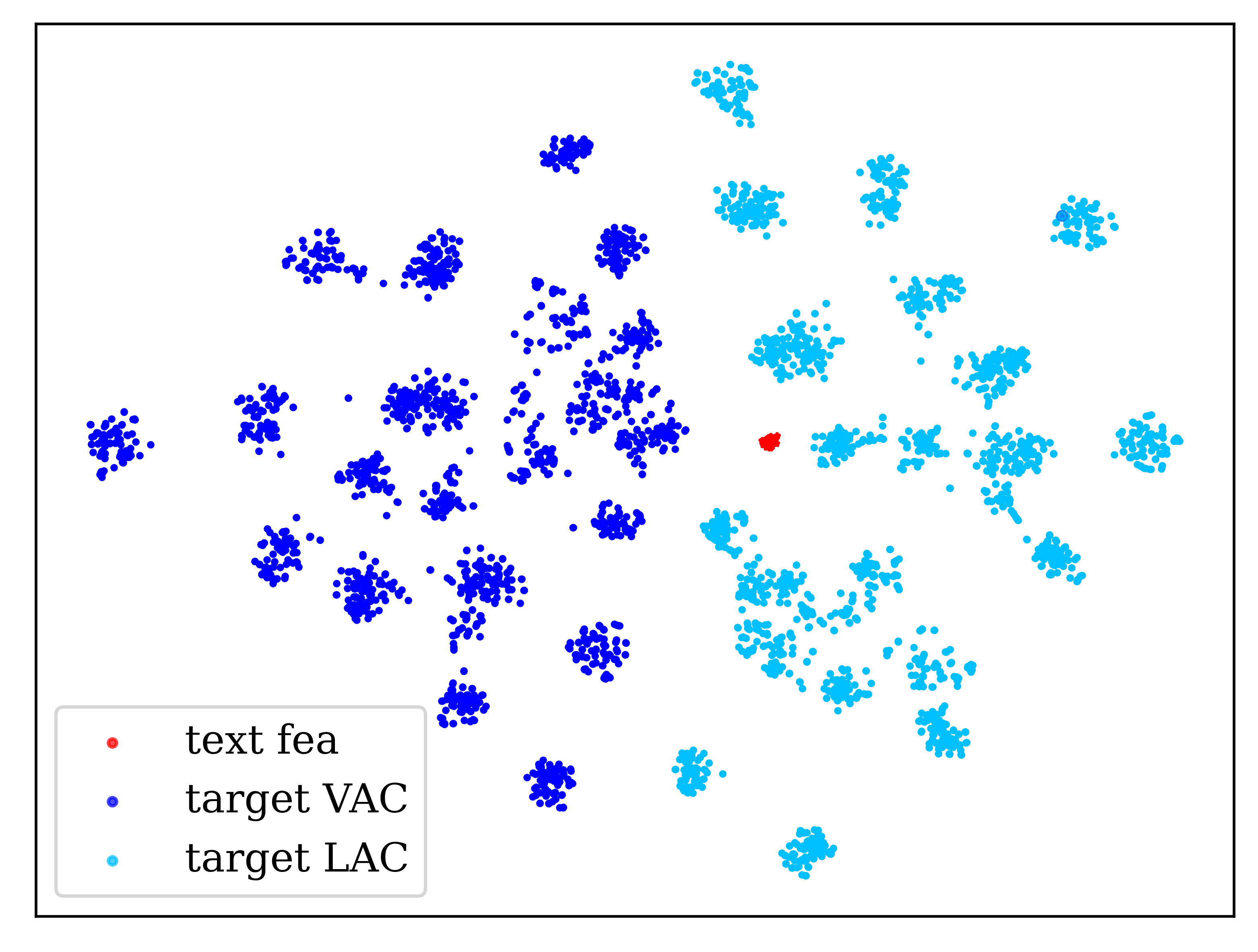}
      \caption{Modality separation effect.}
      \label{tsne_sep}
  \end{subfigure}
  \begin{subfigure}{0.32\textwidth}
    \includegraphics[width=\textwidth]{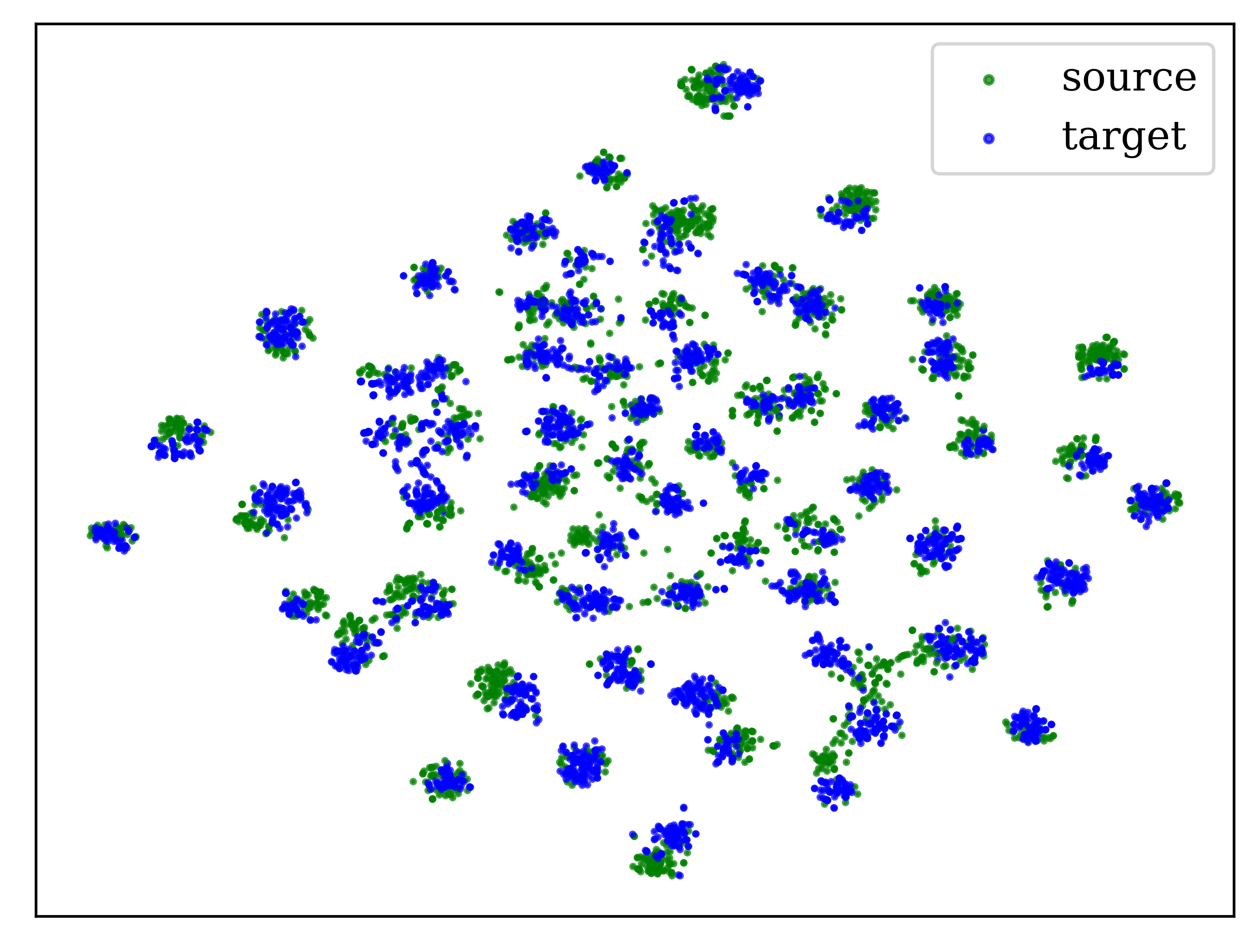}
    \caption{Aligned bottleneck feature distribution.}
    \label{tsne_bott}
  \end{subfigure}
  \vspace{-8pt}
  \caption{T-sne visualization~\cite{van2008visualizing} of the effects of UniMoS on A$\to$P task from Office-Home. UniMoS effectively disentangles CLIP-extracted vision features (\cref{tsne_init}) into LAC and VAC (\cref{tsne_sep}, obtained by randomly selecting 25 classes), and constructs clear cross-domain locality structures (\cref{tsne_bott}).}
  \vspace{-12pt}
  \label{tsne}
\end{figure*}

\bfsection{Computation analysis} \cref{computation} compares computation costs of different methods on VisDA dataset. Our method necessitates training only a few linear layers without updating CLIP backbones, bringing great parameter efficiency. Furthermore, only one forward through CLIP is needed, which allows UniMoS achieve more than \textbf{47$\bm{\times}$} training speed boost than PADCLIP. Prompt learning methods like DAPrompt also requires no training on CLIP backbones, but they  require extensive iterative forwarding of data through CLIP to learn optimal prompts per class, leading to low computing efficiency and scalability to larger datasets. When running on DomainNet with 345 classes, DAPrompt requires more than 22G GPU memory, while our method requires less than 3G. More details are in Supplementary. 

\input{tab/tab_5.tex}

\input{tab/tab_7.tex}

\bfsection{Feature distribution visualization} To demonstrate the efficacy of modality separation and feature alignment, we perform t-sne~\cite{van2008visualizing} visualization on various features at different phases of our method. \cref{tsne_init} shows CLIP-extracted vision features from source (green) and target (blue) domain, along with CLIP-extracted text features (red) of naive prompts. The text features distribute distantly with vision features, proving the existence of modality gap. Besides, CLIP-extracted vision features form poor class discriminability and locality structures. \cref{tsne_sep} showcases the separated LAC and VAC of our method. A clear boundary can be observed between the features of both modalities, indicating the effectiveness of modality separation. Additionally, the text features distribute closer to LAC, proving that the separated LAC is indeed more relevant to the text modality. Following feature alignment and VAC training, the bottleneck features $f_b$ from both domains display a compact class-level locality structure, as demonstrated in \cref{tsne_bott}. This compact structure contributes significantly to the accuracy of the final classification results.

\footnotetext[1]{Cited from PADCLIP~\cite{padclip}. PADCLIP conducts the experiments on an NVIDIA Tesla V100 GPU. Results on other metrics are unavailable since the authors have not released the code implementations yet.}

\begin{figure}[t]
  \centering
  \begin{subfigure}{0.23\textwidth}
      \includegraphics[width=\textwidth]{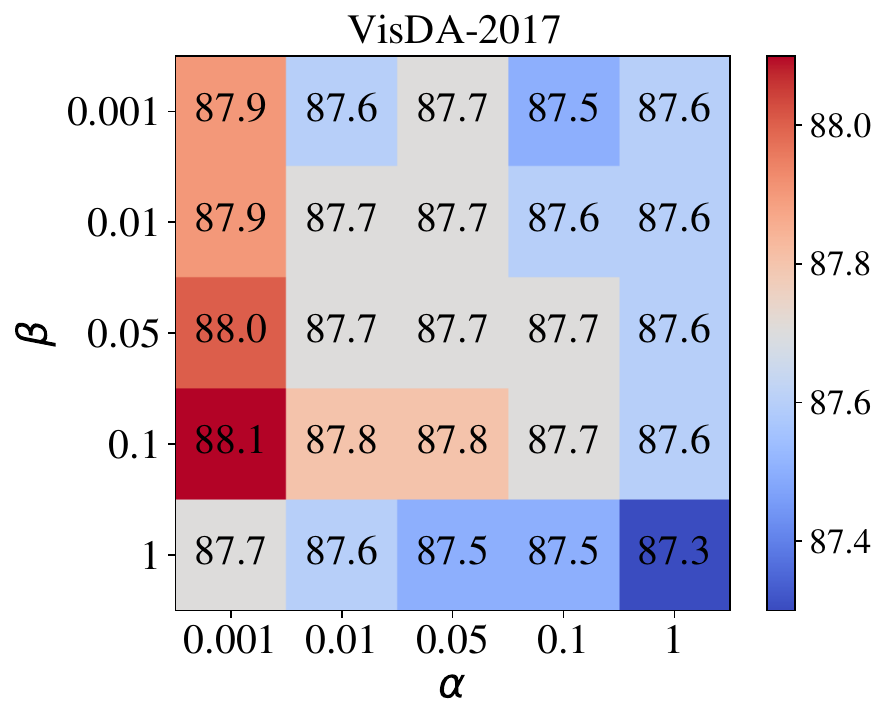}
      \caption{VisDA-2017}
      \label{visda_sen}
  \end{subfigure}
  \begin{subfigure}{0.23\textwidth}
      \includegraphics[width=\textwidth]{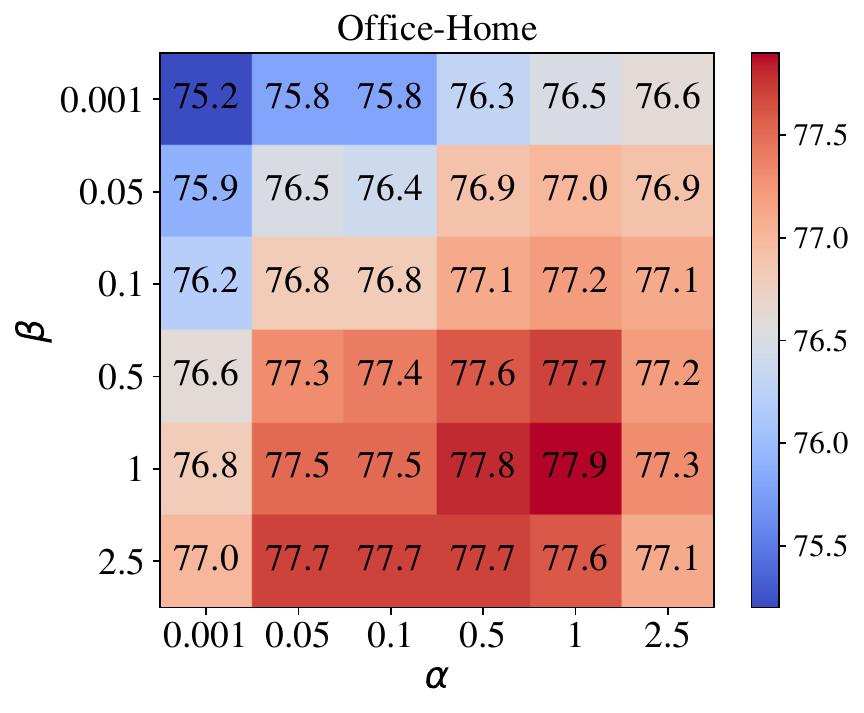}
      \caption{Office-Home}
      \label{officehome_sen}
  \end{subfigure}
  \vspace{-8pt}
  \caption{Parameter sensitivity analysis on $\alpha$ and $\beta$ of UniMoS. }
  \vspace{-12pt}
  \label{sen}
\end{figure}

\bfsection{Hyperparameter sensitivity} In the calibration of UniMoS, we encounter three hyperparameters to determine: $\alpha$ in \cref{llac}, $\beta$ in \cref{lvac}, and $\gamma$ in \cref{opt}.  We empirically discover that alternating $\gamma$ has little impact within each dataset, so we focus on exploring the effects of $\alpha$ and $\beta$. As illustrated in \cref{visda_sen}, the performance on VisDA is notably influenced by $\alpha$, with smaller values leading to improved accuracy. This outcome is attributed to the fact, as discussed in \cref{main_result}, that CLIP struggles to adequately identify source synthetic images from VisDA. Consequently, down-weighting source supervision on LAC  proves beneficial. Conversely, source supervision from both modalities of Office-Home are important. They positively and equally contributes to the adaptation process, so $\alpha=1$ and $\beta=1$ brings the best result, as shown in \cref{officehome_sen}.

\section{Conclusions}
Inspired by the theory of modality gap, in this paper we propose a Unified Modality Separation framework for unsupervised domain adaptation. The CLIP-extracted vision features are explicitly disentangled into vision-associated  and language-associated components, which are trained differently according to their modality strengths and further aligned by a modality discriminator. A modality-ensemble training paradigm unifies both components to leverage modality-specific information while preserving modality-shared contexts, contributing to successful classification. This work is hope to inspire further analysis and exploitation of the multimodal features in pretrained VLMs.

%% file: tab/tab_1.tex
\begin{table*}[ht]
    \caption{UDA results on Office-Home. Best results are marked in bold font. Methods with `*' are based on CLIP.}
    \vspace{-8pt}
    \centering
    \label{officehome}
    %\scriptsize
    %\begin{tabular}{c|c|c \hhs c \hhs c \hhs c \hhs c \hhs c \hhs c \hhs c \hhs c \hhs c \hhs c \hhs c|c}
    \resizebox{\linewidth}{!}{ 
    \begin{tabular}{c|c|cccccccccccc|c}
    \toprule
    %Method & Backbone & Ar$\to$Cl & Ar$\to$Pr & Ar$\to$Rw & Cl$\to$Ar & Cl$\to$Pr & Cl$\to$Rw & Pr$\to$Ar & Pr$\to$Cl & Pr$\to$Rw & Rw$\to$Ar & Rw$\to$Cl & Rw$\to$Pr & Avg. \\ \midrule
    Method & Backbone & A$\to$C & A$\to$P & A$\to$R & C$\to$A & C$\to$P & C$\to$R & P$\to$A & P$\to$C & P$\to$R & R$\to$A & R$\to$C & R$\to$P & Avg. \\ \midrule
    SourceOnly~\cite{resnet} & \multirow{10}{*}{ResNet50} & 34.9 & 50.0 & 58.0 & 37.4 & 41.9 & 46.2 & 38.5 & 31.2 & 60.4 & 53.9 & 41.2 & 59.9 & 46.1 \\
    ParetoDA~\cite{pareto} &  & 56.8 & 75.9 & 80.5 & 64.4 & 73.5 & 73.7 & 65.6 & 55.2 & 81.3 & 75.2 & 61.1 & 83.9 & 70.6 \\
    SDAT~\cite{sdat} &  & 58.2 & 77.1 & 82.2 & 66.3 & 77.6 & 76.8 & 63.3 & 57.0 & 82.2 & 74.9 & 64.7 & 86.0 & 72.2 \\
    MSGD~\cite{msgd} &  & 58.7 & 76.9 & 78.9 & 70.1 & 76.2 & 76.6 & 69.0 & 57.2 & 82.3 & 74.9 & 62.7 & 84.5 & 72.3 \\
    Fixbi~\cite{fixbi} &  & 58.1 & 77.3 & 80.4 & 67.7 & 79.5 & 78.1 & 65.8 & 57.9 & 81.7 & 76.4 & 62.9 & 86.7 & 72.7 \\
    CST~\cite{cst} &  & 59.0 & 79.6 & 83.4 & 68.4 & 77.1 & 76.7 & 68.9 & 56.4 & 83.0 & 75.3 & 62.2 & 85.1 & 72.9 \\
    ATDOC~\cite{atdoc} &  & 60.2 & 77.8 & 82.2 & 68.5 & 78.6 & 77.9 & 68.4 & 58.4 & 83.1 & 74.8 & 61.5 & 87.2 & 73.2 \\
    KUDA~\cite{kuda} &  & 58.2 & 80.0 & 82.9 & 71.1 & 80.3 & 80.7 & 71.3 & 56.8 & 83.2 & 75.5 & 60.3 & 86.6 & 73.9 \\
    EIDCo~\cite{eidco} &  & \textbf{63.8} & 80.8 & 82.6 & 71.5 & 80.1 & 80.9 & 72.1 & 61.3 & 84.5 & \textbf{78.6} & 65.8 & 87.1 & 75.8 \\
    ICON~\cite{icon} &  & 63.3 & 81.3 & 84.5 & 70.3 & 82.1 & 81.0 & 70.3 & \textbf{61.8} & 83.7 & 75.6 & \textbf{68.6} & 87.3 & 75.8 \\ \midrule
    \textcolor{gray}{PADCLIP*}~\cite{padclip} & {\begin{tabular}[c]{@{}c@{}}\textcolor{gray}{ResNet50-}\\ \textcolor{gray}{full-tuning}\end{tabular}} & \textcolor{gray}{57.5} & \textcolor{gray}{84.0} & \textcolor{gray}{83.8} & \textcolor{gray}{\textbf{77.8}} & \textcolor{gray}{85.5} & \textcolor{gray}{84.7} & \textcolor{gray}{76.3} & \textcolor{gray}{59.2} & \textcolor{gray}{85.4} & \textcolor{gray}{78.1} & \textcolor{gray}{60.2} & \textcolor{gray}{86.7} & \textcolor{gray}{76.6} \\ \midrule
    CLIP*~\cite{clip} & \multirow{4}{*}{\begin{tabular}[c]{@{}c@{}}ResNet50- \\     none-tuning\end{tabular}} & 51.7 & 81.5 & 82.3 & 71.7 & 81.5 & 82.3 & 71.7 & 51.7 & 82.3 & 71.7 & 51.7 & 81.5 & 71.8 \\
    DAPrompt*~\cite{dapl} &  & 54.1 & 84.3 & 84.8 & 74.4 & 83.7 & 85.0 & 74.5 & 54.6 & 84.8 & 75.2 & 54.7 & 83.8 & 74.5 \\
    ADCLIP*~\cite{adclip} &  & 55.4 & 85.2 & 85.6 & 76.1 & 85.8 & 86.2 & \textbf{76.7} & 56.1 & 85.4 & 76.8 & 56.1 & 85.5 & 75.9 \\
    \rowcolor[HTML]{d1cfcf} 
    \textbf{UniMoS* (ours)} &  & 59.5 & \textbf{89.4} & \textbf{86.9} & 75.2 & \textbf{89.6} & \textbf{86.8} & 75.4 & 58.4 & \textbf{87.2} & 76.9 & 59.5 & \textbf{89.7} & \textbf{77.9} \\ \bottomrule
    %PADCLIP*~\cite{padclip} & {\begin{tabular}[c]{@{}c@{}}ResNet50-\\      tuning\end{tabular}} & 57.5 & 84.0 & 83.8 & \textbf{77.8} & 85.5 & 84.7 & 76.3 & \textbf{59.2} & 85.4 & 78.1 & 60.2 & 86.7 & 76.6 \\ \bottomrule

    % padclip放上面去
    
    %CDTrans~\cite{cdtrans} & DeiT-B & 68.8 & 85.0 & 86.9 & 81.5 & 87.1 & 87.3 & 79.6 & 63.3 & 88.2 & 82.0 & 66.0 & 90.6 & 80.5 \\ \midrule
    %SourceOnly~\cite{vit} & \multirow{8}{*}{ViT-B} & 54.7 & 83.0 & 87.2 & 77.3 & 83.4 & 85.5 & 74.4 & 50.9 & 87.2 & 79.6 & 53.8 & 88.8 & 75.5 \\
    %CLIP*~\cite{clip} &  & 67.8 & 89.0 & 89.8 & 82.9 & 89.0 & 89.8 & 82.9 & 67.8 & 89.8 & 82.9 & 67.8 & 89.0 & 82.4 \\
    %TVT~\cite{tvt} &  & 74.9 & 86.8 & 89.5 & 82.8 & 88.0 & 88.3 & 79.8 & 71.9 & 90.1 & 85.5 & 74.6 & 90.6 & 83.6 \\
    %DAPrompt*~\cite{dapl} &  & 70.6 & 90.2 & 91.0 & 84.9 & 89.2 & 90.9 & 84.8 & 70.5 & 90.6 & 84.8 & 70.1 & 90.8 & 84.0 \\
    %SSRT~\cite{ssrt} &  & 75.2 & 89.0 & 91.1 & 85.1 & 88.3 & 89.9 & 85.0 & 74.2 & 91.2 & 85.7 & 78.6 & 91.8 & 85.4 \\
    %ADCLIP*~\cite{adclip} &  & 70.9 & 92.5 & 92.1 & 85.4 & 92.4 & 92.5 & \textbf{86.7} & 74.3 & \textbf{93.0} & \textbf{86.9} & 72.6 & 93.8 & 86.1 \\
    %PADCLIP*~\cite{padclip} &  & \textbf{76.4} & 90.6 & 90.8 & \textbf{86.7} & 92.3 & 92.0 & 86.0 & \textbf{74.5} & 91.5 & \textbf{86.9} & \textbf{79.1} & 93.1 & 86.7 \\
    %\rowcolor[HTML]{d1cfcf} 
    %UniMoS* (ours) &  & 74.9 & \textbf{94.0} & \textbf{92.5} & 86.4 & \textbf{94.3} & \textbf{92.5} & 86.0 & 73.9 & \textbf{93.0} & 86.4 & 74.2 & \textbf{94.5} & \textbf{86.9} \\ \bottomrule
    \end{tabular}}
    \vspace{-4pt}
\end{table*}

%% file: tab/tab_2.tex
\begin{table*}[ht]
    \caption{UDA results on VisDA-2017. Best results are marked in bold font. Methods with `*' are based on CLIP.}
    \vspace{-8pt}
    \label{visda}
    \footnotesize
    \centering
    \resizebox{\linewidth}{!}{
    \begin{tabular}{c|c|cccccccccccc|c}
    \toprule
    Method & Backbone & plane & bicycle & bus & car & horse & knife & mcycl & person & plant & sktbrd & train & truck & Avg. \\ \midrule
    SourceOnly~\cite{resnet} & \multirow{6}{*}{ResNet101}  & 55.1 & 53.3 & 61.9 & 59.1 & 80.6 & 17.9 & 79.7 & 31.2 & 81.0 & 26.5 & 73.5 & 8.5 & 52.4 \\
    ParetoDA~\cite{pareto} &  & 95.9 & 82.8 & 81.3 & 58.7 & 93.9 & 93.7 & 85.9 & 83.0 & 91.9 & 92.0 & 87.1 & 51.8 & 83.2 \\
    MSGD~\cite{msgd} &  & 97.5 & 83.4 & 84.4 & 69.4 & 95.9 & 94.1 & 90.9 & 75.5 & 95.5 & 94.6 & 88.1 & 44.9 & 84.5 \\
    ATDOC~\cite{atdoc} &  & 95.3 & 84.7 & 82.4 & 75.6 & 95.8 & \textbf{97.7} & 88.7 & 76.6 & 94.0 & 91.7 & 91.5 & 61.9 & 86.3 \\
    CAN~\cite{can} &  & 97.0 & 87.2 & 82.5 & 74.3 & 97.8 & 96.2 & 90.8 & 80.7 & 96.6 & \textbf{96.3} & 87.5 & 59.9 & 87.2 \\
    FixBi~\cite{fixbi} &  & 96.1 & 87.8 & 90.5 & \textbf{90.3} & 96.8 & 95.3 & 92.8 & \textbf{88.7} & \textbf{97.2} & 94.2 & 90.9 & 25.7 & 87.2 \\ \midrule
    \textcolor{gray}{PADCLIP*}~\cite{padclip} & {\begin{tabular}[c]{@{}c@{}}\textcolor{gray}{ResNet101-}\\ \textcolor{gray}{full-tuning}\end{tabular}} & \textcolor{gray}{96.7} & \textcolor{gray}{\textbf{88.8}} & \textcolor{gray}{87.0} & \textcolor{gray}{82.8} & \textcolor{gray}{97.1} & \textcolor{gray}{93.0} & \textcolor{gray}{91.3} & \textcolor{gray}{83.0} & \textcolor{gray}{95.5} & \textcolor{gray}{91.8} & \textcolor{gray}{91.5} & \textcolor{gray}{63.0} & \textcolor{gray}{\textbf{88.5}} \\ \midrule
    CLIP*~\cite{clip} & \multirow{4}{*}{\begin{tabular}[c]{@{}c@{}}ResNet101-\\   none-tuning\end{tabular}} & \textbf{98.2} & 83.9 & 90.5 & 73.5 & 97.2 & 84.0 & 95.3 & 65.7 & 79.4 & 89.9 & 91.8 & 63.3 & 84.4 \\
    DAPrompt*~\cite{dapl} &  & 97.8 & 83.1 & 88.8 & 77.9 & 97.4 & 91.5 & 94.2 & 79.7 & 88.6 & 89.3 & \textbf{92.5} & 62.0 & 86.9 \\
    ADCLIP*~\cite{adclip} &  & 98.1 & 83.6 & \textbf{91.2} & 76.6 & \textbf{98.1} & 93.4 & \textbf{96.0} & 81.4 & 86.4 & 91.5 & 92.1 & 64.2 & 87.7 \\
    \rowcolor[HTML]{d1cfcf} 
    \textbf{UniMoS* (ours)} &  & 97.7 & 88.2 & 90.1 & 74.6 & 96.8 & 95.8 & 92.4 & 84.1 & 90.8 & 89.0 & 91.8 & \textbf{65.3} & 88.1 \\ \bottomrule
    %PADCLIP*~\cite{padclip} & {\begin{tabular}[c]{@{}c@{}}ResNet101-\\ tuning\end{tabular}} & 96.7 & \textbf{88.8} & 87.0 & 82.8 & 97.1 & 93.0 & 91.3 & 83.0 & 95.5 & 91.8 & 91.5 & 63.0 & \textbf{88.5} \\  \bottomrule
    \end{tabular}}
    \vspace{-8pt}
\end{table*}

%% file: tab/tab_3.tex
\begin{table*}[ht]
    \caption{UDA results on DomainNet. Best results are marked in bold font. Methods with `*' are based on CLIP.}
    \vspace{-8pt}
    \label{domainnet}
    \resizebox{\linewidth}{!}{ 
    \begin{tabular}{@{}c|ccccccc||c|ccccccc||c|ccccccc}
    \hline
    \begin{tabular}[c]{@{}c@{}}DeiT\\ -B~\cite{deit}\end{tabular} & clp & inf & pnt & qdr & rel & skt & avg & \begin{tabular}[c]{@{}c@{}}ViT\\ -B~\cite{vit}\end{tabular} & clp & inf & pnt & qdr & rel & skt & avg & \begin{tabular}[c]{@{}c@{}}SSRT\\ -B~\cite{ssrt}\end{tabular} & clp & inf & pnt & qdr & rel & skt & avg \\ 
    \hline
    clp & - & 24.3 & 49.6 & 15.8 & 65.3 & 52.1 & 41.4 & clp & - & 27.2 & 53.1 & 13.2 & 71.2 & 53.3 & 43.6 & clp &  & 33.8 & 60.2 & 19.4 & 75.8 & 59.8 & 49.8 \\ 
    inf & 45.9 & - & 45.9 & 6.7 & 61.4 & 39.5 & 39.9 & inf & 51.4 & - & 49.3 & 4.0 & 66.3 & 41.1 & 42.4 & inf & 55.5 & - & 54.0 & 9.0 & 68.2 & 44.7 & 46.3 \\ 
    pnt & 53.2 & 23.8 & - & 6.5 & 66.4 & 44.7 & 38.9 & pnt & 53.1 & 25.6 & - & 4.8 & 70.0 & 41.8 & 39.1 & pnt & 61.7 & 28.5 & - & 8.4 & 71.4 & 55.2 & 45.0 \\ 
    qdr & 31.9 & 6.8 & 15.4 & - & 23.4 & 20.6 & 19.6 & qdr & 30.5 & 4.5 & 16.0 & - & 27.0 & 19.3 & 19.5 & qdr & 42.5 & 8.8 & 24.2 & - & 37.6 & 33.6 & 29.3 \\ 
    rel & 59.0 & 25.8 & 56.3 & 9.2 & - & 44.8 & 39.0 & rel & 58.4 & 29.0 & 60.0 & 6.0 & - & 45.8 & 39.9 & rel & 69.9 & 37.1 & 66.0 & 10.1 & - & 58.9 & 48.4 \\ 
    skt & 60.6 & 20.6 & 48.4 & 16.5 & 61.2 & - & 41.5 & skt & 63.9 & 23.8 & 52.3 & 14.4 & 67.4 & - & 44.4 & skt & 70.6 & 32.8 & 62.2 & 21.7 & 73.2 & - & 52.1 \\ 
    avg & 50.1 & 20.3 & 43.1 & 10.9 & 55.5 & 40.3 & \colorbox{lightgray}{36.7} & avg & 51.5 & 22.0 & 46.1 & 8.5 & 60.4 & 40.3 & \colorbox{lightgray}{38.1} & avg & 60.0 & 28.2 & 53.3 & 13.7 & 65.3 & 50.4 & \colorbox{lightgray}{45.2} \\ \hline
    \begin{tabular}[c]{@{}c@{}}CDTrans\\ -DeiT~\cite{cdtrans}\end{tabular} & clp & inf & pnt & qdr & rel & skt & avg & \begin{tabular}[c]{@{}c@{}}PMTrans\\ -Swin~\cite{pmtrans}\end{tabular} & clp & inf & pnt & qdr & rel & skt & avg & \begin{tabular}[c]{@{}c@{}}CLIP\\ -B*~\cite{clip}\end{tabular} & clp & inf & pnt & qdr & rel & skt & avg \\ \hline
    clp & - & 29.4 & 57.2 & 26.0 & 72.6 & 58.1 & 48.7 & clp & - & 34.2 & 62.7 & 32.5 & 79.3 & 63.7 & 54.5 & clp & - & 70.1 & 70.1 & 70.1 & 70.1 & 70.1 & 70.1 \\
    inf & 57.0 & - & 54.4 & 12.8 & 69.5 & 48.4 & 48.4 & inf & 67.4 & - & 61.1 & 22.2 & 78.0 & 57.6 & \textbf{57.3} & inf & 46.4 & - & 46.4 & 46.4 & 46.4 & 46.4 & 46.4 \\ 
    pnt & 62.9 & 27.4 & - & 15.8 & 72.1 & 53.9 & 46.4 & pnt & 69.7 & 33.5 & - & 23.9 & 79.8 & 61.2 & 53.6 & pnt & 61.7 & 61.7 & - & 61.7 & 61.7 & 61.7 & 61.7 \\ 
    qdr & 44.6 & 8.9 & 29.0 & - & 42.6 & 28.5 & 30.7 & qdr & 54.6 & 17.4 & 38.9 & - & 49.5 & 41.0 & \textbf{40.3} & qdr & 13.7 & 13.7 & 13.7 & - & 13.7 & 13.7 & 13.7 \\ 
    rel & 66.2 & 31.0 & 61.5 & 16.2 & - & 52.9 & 45.6 & rel & 74.1 & 35.3 & 70.0 & 25.4 & - & 61.1 & 53.2 & rel & 82.9 & 82.9 & 82.9 & 82.9 & - & 82.9 & 82.9 \\ 
    skt & 69.0 & 29.6 & 59.0 & 27.2 & 72.5 & - & 51.5 & skt & 73.8 & 33.0 & 62.6 & 30.9 & 77.5 & - & 55.6 & skt & 62.6 & 62.6 & 62.6 & 62.6 & 62.6 & - & 62.6 \\ 
    avg & 59.9 & 25.3 & 52.2 & 19.6 & 65.9 & 48.4 & \colorbox{lightgray}{45.2} & avg & 67.9 & 30.7 & 59.1 & 27.0 & 72.8 & 56.9 & \colorbox{lightgray}{52.4} & avg & 53.5 & 58.2 & 55.1 & 64.7 & 50.9 & 55.0 & \colorbox{lightgray}{56.2} \\ \hline
    \begin{tabular}[c]{@{}c@{}}DAPrompt\\ -B*~\cite{dapl}\end{tabular} & clp & inf & pnt & qdr & rel & skt & avg & \begin{tabular}[c]{@{}c@{}}\textcolor{gray}{PADCLIP}\\ \textcolor{gray}{-B*}~\cite{padclip}\end{tabular} & clp & inf & pnt & qdr & rel & skt & avg & \begin{tabular}[c]{@{}c@{}}\textbf{UniMoS}\\ \textbf{-B*} (ours)\end{tabular} & clp & inf & pnt & qdr & rel & skt & avg \\ \hline
    clp & - & 73.0 & 73.8 & 72.6 & 73.9 & 73.5 & 73.4 & clp & - & 73.6 & 75.4 & 74.6 & 76.4 & 76.3 & 75.3 & clp & - & 76.5 & 77.2 & 76.6 & 77.5 & 77.8 & \textbf{77.1} \\ 
    inf & 50.8 & - & 50.1 & 49.6 & 50.6 & 50.3 & 50.3 & inf & 55.1 & - & 54.3 & 53.6 & 54.9 & 54.9 & 54.6 & inf & 55.1 & - & 55.0 & 54.6 & 55.3 & 55.2 & 55.0 \\ 
    pnt & 70.2 & 69.6 & - & 68.9 & 70.4 & 69.9 & 69.8 & pnt & 71.1 & 70.6 & - & 70.0 & 72.7 & 71.7 & 71.2 & pnt & 72.3 & 71.5 & - & 69.4 & 72.5 & 72.6 & \textbf{71.7} \\ 
    qdr & 17.2 & 14.4 & 13.9 & - & 14.3 & 13.9 & 14.7 & qdr & 36.8 & 18.0 & 32.0 & - & 31.7 & 34.9 & 30.7 & qdr & 25.0 & 22.9 & 23.6 & - & 23.7 & 25.1 & 24.1 \\ 
    rel & 84.9 & 84.8 & 84.9 & 84.7 & - & 84.6 & 84.8 & rel & 84.2 & 83.5 & 83.5 & 83.1 & - & 83.6 & 83.6 & rel & 86.0 & 85.9 & 85.8 & 85.5 & - & 85.9 & \textbf{85.8} \\ 
    skt & 65.8 & 65.4 & 65.8 & 64.9 & 65.9 & - & 65.6 & skt & 68.1 & 66.6 & 67.2 & 66.1 & 67.5 & - & 67.1 & skt & 68.5 & 67.8 & 68.2 & 67.5 & 68.0 & - & \textbf{68.0} \\ 
    avg & 57.8 & 61.4 & 57.7 & 68.1 & 55.0 & 58.4 & \colorbox{lightgray}{59.8} & avg & 63.1 & 62.5 & 62.5 & 69.5 & 60.6 & 64.3 & \colorbox{lightgray}{\textbf{63.7}} & avg & 61.4 & 64.9 & 62.0 & 70.7 & 59.4 & 63.3 & \colorbox{lightgray}{63.6} \\ \hline
    \end{tabular}}
    \vspace{-8pt}
\end{table*}

%% file: tab/tab_4.tex
\begin{table}[t]
    \centering
    \caption{UDA results on Mini-DomainNet. Best results are marked in bold font. All compared methods are CLIP-based.}
    \vspace{-5pt}
    \label{mini_domainnet}
    \resizebox{\linewidth}{!}{
      \begin{tabular}{c|c|c|c|c|c}
        \toprule
        Method & Backbone & Acc. & Method & Backbone & Acc. \\ \midrule
        %SourceOnly & \multirow{5}{*}{ResNet50} &  & SourceOnly & \multirow{5}{*}{ViT-B} &  \\
        CLIP & \multirow{4}{*}{ResNet50} & 71.2 & CLIP & \multirow{4}{*}{ViT-B} & 82.8 \\
        DAPrompt &  & 74.8 & DAPrompt &  & 85.8 \\
        ADCLIP &  & 75.2 & ADCLIP &  & 86.9 \\
        \rowcolor[HTML]{d1cfcf}
        \textbf{UniMoS (ours)} &  & \textbf{78.0} & \textbf{UniMoS (ours)} &  & \textbf{87.3} \\ \bottomrule
      \end{tabular}}
    \vspace{-5pt}
\end{table}

%% file: tab/tab_6.tex
\begin{table}[t]
  \caption{Ablation study on Office-Home and VisDA-2017.}
  \vspace{-5pt}
  \centering
  \label{ablation}
  \resizebox{0.88\linewidth}{!}{
  \begin{tabular}{c|cc}
  \toprule
  Method & Office-Home & VisDA-2017 \\ \midrule
  w/o $\mathcal{L}_{ortho}$ & 77.4 & 87.6 \\
  w/o debiasing & 77.3 & 87.7 \\
  w/o $\mathcal{L}_{im}$ & 77.0 & 87.8 \\
  w/o $\mathcal{L}_{distill}$ & 77.0 & 87.6 \\
  w/o learnable weight & 76.9 & 86.2 \\
  %\begin{tabular}[c]{@{}c@{}}w/o   modality\\      Discriminator\end{tabular} & 77.6 & 87.9 \\
  w/o  modality  discriminator & 77.6 & 87.9 \\
  \textbf{UniMoS (full design)} & \textbf{77.9} & \textbf{88.1} \\ \bottomrule
  \end{tabular}}
  \vspace{-12pt}
\end{table}

%% file: tab/tab_5.tex
\begin{table}[t]
  \caption{Results using different backbones on Office-Home. Best results are marked in bold font. All compared methods are based on CLIP.}
  \vspace{-5pt}
  \centering
  \label{backbones}
  \resizebox{1\linewidth}{!}{
    \begin{tabular}{c|c|c|c|c|c}
      \toprule
      Method & Backbone & Acc. & Method & Backbone & Acc. \\ \midrule
      \textbf{UniMoS (ours)} & ResNet50 & \textbf{77.9} & CLIP & \multirow{5}{*}{ViT-B} & 82.4 \\ \cmidrule{1-3}
      CLIP & \multirow{4}{*}{ViT-L} & 87.0 & DAPrompt &  & 84.4 \\
      DAPrompt &  & 88.7 & ADCLIP &  & 86.1 \\
      ADCLIP &  & 90.5 & \textcolor{gray}{PADCLIP} &  & \textcolor{gray}{86.7} \\
      \rowcolor[HTML]{d1cfcf}
      \textbf{UniMoS (ours)} &  & \textbf{90.7} & \textbf{UniMoS (ours)} &  & \textbf{86.9} \\ \bottomrule
    \end{tabular}}
  \vspace{-5pt}
\end{table}

%% file: tab/tab_7.tex
\begin{table}[t]
    \caption{Computation analysis on VisDA-2017. Best results are marked in bold font. Methods with `*' are based on CLIP.}
    \vspace{-5pt}
    \label{computation}
    \resizebox{\linewidth}{!}{
      \begin{tabular}{c|c|cccc|c}
        \toprule
    Method & Bkb. & Param. & \begin{tabular}[c]{@{}c@{}}Throughput\\      (imges/s)\end{tabular} & \begin{tabular}[c]{@{}c@{}}Train\\      time\end{tabular} & FLOPs & Acc. \\ \midrule
    DAPrompt* &  & 1.2M & 244 & 4.3H & 11.3G & 86.9 \\
    FixBi &  & 86.1M & 102 & 5.5H & 15.73G & 87.2 \\
    CAN &  & 42.5M & 31 & 10.5H & 7.9G & 87.2 \\ 
    \textcolor{gray}{PADCLIP*} &  & - & - & \textcolor{gray}{23.5H}\footnotemark[1]  & - & \textcolor{gray}{\textbf{88.5}} \\
    \rowcolor[HTML]{d1cfcf}
    \textbf{UniMoS* (ours)} & \multirow{-5}{*}{\rotatebox{90}{ResNet101}} & \textbf{0.79M} & \textbf{2667} & \textbf{0.5H} & \textbf{\textless{}0.01G} & 88.1 \\ \bottomrule 
    \end{tabular}}
    \vspace{-8pt}
\end{table}

%% file: sec/X_suppl.tex
\maketitlesupplementary

\begin{algorithm}[t]
    \caption{Training Algorithm of UniMoS}
    \label{alg1}
    \textbf{Input}: Labeled source data $\{x^s, y^s\}$, unlabeled target data $x^t$, maximum epoch number $max\_epoch$, pretrained CLIP text encoder $g_{txt}$ and vision encoder $g_{vis}$.\\
    \textbf{Output}: Trained modality separator $G_{txt},G_{vis}$, and trained linear layers $\Phi_1,\Phi_2$.
    \begin{algorithmic}[1] %[1] enables line numbers
    \STATE Let $epoch=0$.
    \WHILE{$epoch < max\_epoch$}
    \STATE Obtain teacher output $y_{lac}^t$ via \cref{lac_sup}.
    \IF {$epoch\mod2==0$}
    \STATE Take mixed outputs as target vision pseudo label via \cref{mixup_sup}.
    \ELSE 
    \STATE Obtain target vision pseudo label $y_{vac}^t$ via \cref{cluster_sup}.
    \ENDIF
    %\STATE Randomly reset half of the parameters of $W$.
    \STATE Obtain CLIP-extracted vision features $\{f_v^s, f_v^t\}=g_{vis}(x^s,x^t)$ and text features $\mu_i=g_{txt}(t_i)$.
    \STATE Obtain LAC outputs $\hat{y}_{lac}$ by \cref{lac_hat_sup}.
    \STATE Obtain VAC outputs $\hat{y}_{vac}$ by \cref{vac_hat_sup}.
    \STATE Obtain ensemble outputs $\hat{y}_{ens}$ by \cref{ensemble_sup}.
    \STATE Obtain modality discrimination outputs $y_{dis}$ by \cref{dis_sup}.
    \STATE Update network parameters via \cref{opt_sup}.
    \STATE Let $epoch=epoch+1$
    \ENDWHILE
    \STATE \textbf{return} solution
    \end{algorithmic}
\label{algorithm_sup}
\end{algorithm}

\section{Training algorithm}
\label{training_sup}
For clarity, we give full training algorithm of UniMoS. We first obtain teacher output from CLIP's zero-shot inference results by \cref{lac_sup}:
\begin{equation}
    y_{lac}^t = (l_1-\overline{l} , l_2-\overline{l}, \cdots, l_k-\overline{l}), \quad l_i = \cos(\mu_{i}, f_{v}^t)/T.
    \label{lac_sup}
\end{equation}
Then obtain pseudo label by \cref{cluster_sup}:
\begin{equation}
    \hat{y}_{vac}^t = \arg \underset{k}{\max} \, \cos(f_b^t, \phi_k).
    \label{cluster_sup}
\end{equation}
To avoid potential error accumulations~\cite{decorate} in pseudo labels, we directly apply mixed outputs from both modalities as pseudo labels in certain epochs:
\begin{equation}
    \hat{y}_{vac}^t = \lambda \cdot \hat{y}_{vac}^t + (1-\lambda) \cdot \tilde{y}_{lac}^t,
    \label{mixup_sup}
\end{equation}
where $\lambda$ is a fixed mixup ratio 0.3 that combines outputs from both modalities for inference. LAC and VAC outputs are obtained via \cref{lac_hat_sup} and \cref{vac_hat_sup}, respectively.
\begin{equation}
    \hat{y}_{lac} = (\hat{l}_1, \hat{l}_2, \cdots \hat{l}_k), \quad \hat{l}_i = \cos(\mu_{i}, f_{lac})/T.
    \label{lac_hat_sup}
\end{equation}
\begin{equation}
    \quad \hat{y}_{vac} = \Phi_2 (f_b), \quad f_b = \Phi_1 (f_{vac}).
    \label{vac_hat_sup}
\end{equation}
We further train a modality discriminator to align modality features from both domains via \cref{dis_sup}:
\begin{equation}
    \mathcal{L}_{bce} = -[y_{dis} \log \hat{y}_{dis} + (1-y_{dis}) \log (1-\hat{y}_{dis})],
    \label{dis_sup}
\end{equation}
We assemble LAC and VAC during training to facilitate modality information exchanges by \cref{ensemble_sup}:
\begin{equation}
    \hat{y}_{ens}^t = w \cdot \hat{y}_{vac}^t + (1-w) \cdot \tilde{y}_{lac}^t.
    \label{ensemble_sup}
\end{equation}
Note that the $w$ in \cref{ensemble_sup} is trainable, which dynamically changes to suit different training phases. While the $\lambda$ in \cref{mixup_sup} is fixed and only used for retrieving pseudo labels and inference, thus does not affect training.
%To provide more flexible ensemble strategies, we opt to randomly reset a part of the parameters of weight generator, as in \cref{algorithm}.
Based on the training objectives:
\begin{align}
    & \theta_{G_{txt}} = \underset{\theta_{G_{txt}}}{\arg \min} \; \mathcal{L}_{lac} + \gamma\mathcal{L}_{ortho} + \gamma\mathcal{L}_{bce}, \label{opt_sup}\\
    & \theta_{G_{vis}} = \underset{\theta_{G_{vis}}}{\arg  \min} \; \mathcal{L}_{vac} + \gamma\mathcal{L}_{ortho} + \gamma\mathcal{L}_{bce}, \nonumber \\
    & \theta_{W}, \, \theta_{\Phi_1}, \, \theta_{\Phi_2} = \underset{\theta_{W},\theta_{\Phi_1},\theta_{\Phi_2}}{\arg \min} \; \mathcal{L}_{vac}, \nonumber \\
    & \theta_{D} = \underset{\theta_{D}}{\arg \min} \; \gamma \mathcal{L}_{bce} \nonumber ,
\end{align}
we update the trainable parameters. Training algorithm is given in \cref{algorithm_sup}. Please refer to main paper for full descriptions of the equations above.

\input{tab/sup_tab_1.tex}

\input{tab/sup_tab_2.tex}

All trainable modules in UniMoS are fully-connected linear layers, which brings very low computation costs. Specifically, the modality separators $G_{txt},G_{vis}$ are two separate linear layers of shape ($d_v,d_v$). The bottleneck feature dimension $d_b$ is 256, thus $\Phi_1$ is of shape ($d_v$,256) with batch normalization, and $\Phi_2$ is of shape (256,$K$). The modality discriminator $D$ consists of two linear layers ($d_v$,256) and (256,1) with a ReLU activation layer. The weight generator $W$ consists of two linear layers ($d_v$,256) and (256,1) with a Sigmoid activation layer.

\section{Additional experiments}
In this section we provide full results of our experiments.

\bfsection{Mini-DomainNet} We present full results on Mini-DomainNet~\cite{saito2019semi,litrico2023guiding,zhang2023rethinking}  in \cref{mini_domainnet_sup} using two backbones. We set $\alpha=0.1$, $\beta=1$ and $\gamma=0.01$ for all tasks. We can observe significant improvements over current SOTA ADCLIP~\cite{adclip} by 2.8\% in average with ResNet50~\cite{resnet}, and 0.4\% improvement using ViT~\cite{vit}. We are unable to compare with PADCLIP~\cite{padclip} since no public code implementation is available yet.

\bfsection{Office-Home} We provide full results on Office-Home~\cite{officehome} in \cref{officehome_sup} using ResNet50 and two variants of ViT as backbone. Utilizing the rich pretrain knowledge in CLIP, CLIP-based methods achieve higher accuracies than single-modality UDA methods. Our method further consistently outperforms all competing methods on all backbones, demonstrating the efficacy of adapting both modalities.

\begin{figure*}[t]
    \centering
    \begin{subfigure}{0.32\textwidth}
        \includegraphics[width=\textwidth]{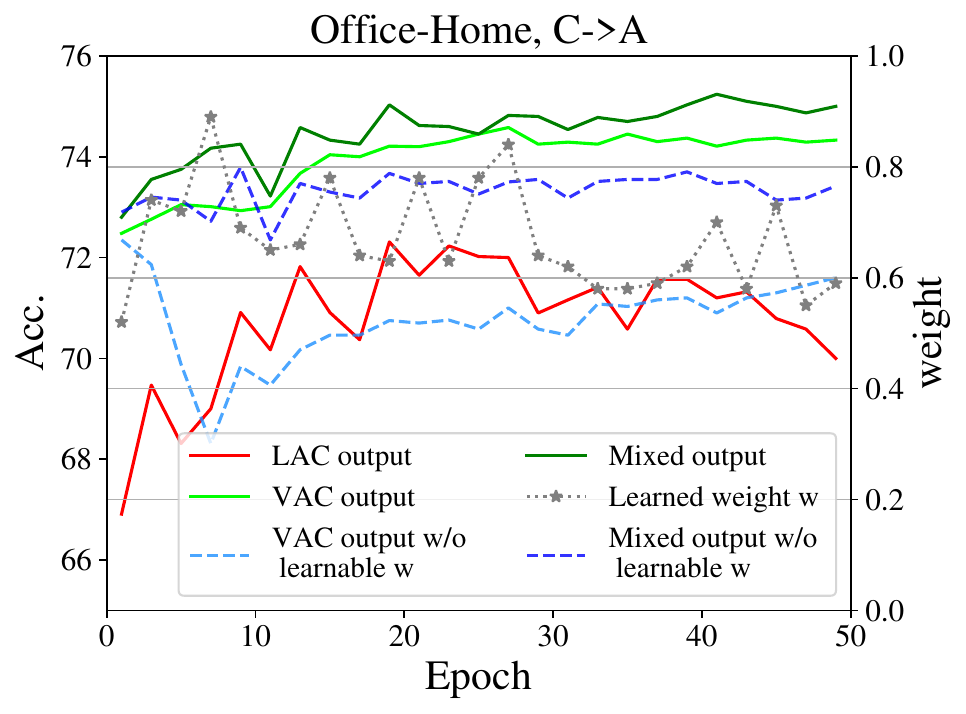}
        \caption{Task C$\to$A}
        %\label{tsne_init}
    \end{subfigure}
    \begin{subfigure}{0.32\textwidth}
        \includegraphics[width=\textwidth]{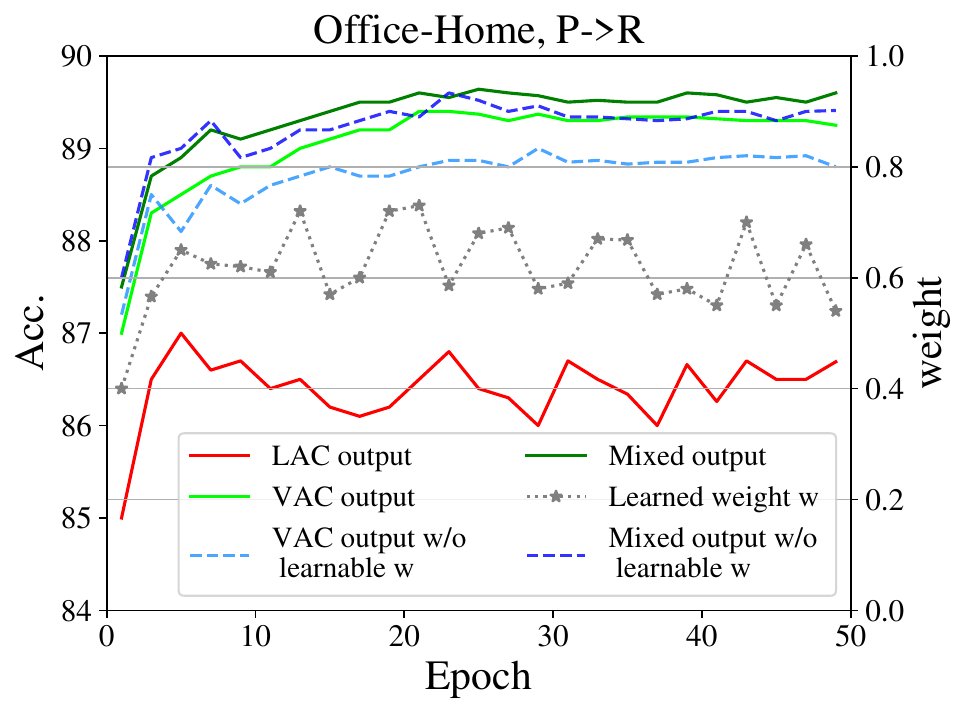}
        \caption{Task P$\to$R}
        \label{w2}
    \end{subfigure}
    \begin{subfigure}{0.32\textwidth}
        \includegraphics[width=\textwidth]{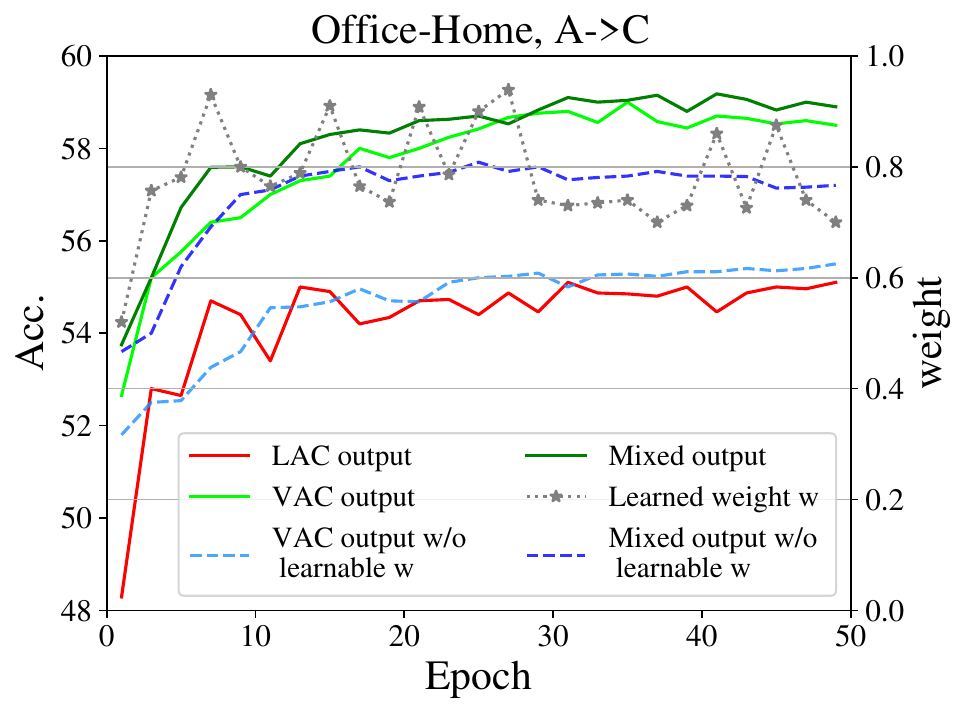}
        \caption{Task A$\to$C}
        %\label{tsne_bott}
    \end{subfigure}
    \vspace{-8pt}
    \caption{Examples on the effects of ensemble weight $w$.}
    \label{w}
\end{figure*}

\input{tab/sup_tab_3.tex}

\bfsection{Computation analysis} We provide comprehensive computation analysis in \cref{computation_sup}.  The GPU memory consumption is computed using consistent batch size of 32. The results on DomainNet is computed on task `clp$\to$inf'. Note our method requires only one forward through CLIP to obtain pre-extracted features, and we have included the time for pre-extraction in the `Train time' columns. According to \cref{training_sup}, only a few linear layers is trained in our method, which brings significant computation efficiency, allowing agile applications of our method. We can observe that the trainable parameters of DAPrompt differs on different datasets. The reason is that DAPrompt trains the prompt embedding for each class respectively,  therefore the trainable parameters and GPU consumption increase drastically as class number increases from 12 (VisDA) to 345 (DomainNet).

\bfsection{Learnable ensemble weight $w$} \cref{w} provides additional examples on the effects of dynamic ensemble weight on three tasks from Office-Home dataset. We can draw the conclusion that the accuracy of VAC drops drastically without learning ensemble weight $w$, which further affects the overall mixed output accuracy. We can observe in \cref{w2} that although accuracy of `VAC output w/o learnable w' is lower than the full design, the final `Mixed output w/o learnable w' achieves comparable accuracy with the full design. The reason is that the task P$\to$R is relatively easier, thus complementary knowledge in LAC is able to support VAC for classification.

%% file: tab/sup_tab_1.tex
\begin{table*}[t]
    \caption{Results on Mini-DomainNet. Best results are marked in bold font. Methods with `*' are based on CLIP.}
    \label{mini_domainnet_sup}
    \resizebox{\linewidth}{!}{
    \begin{tabular}{c|c|cccccccccccc|c}
    \toprule
    Method & Backbone & clp$\to$pnt & clp$\to$rel & clp$\to$skt & pnt$\to$clp & pnt$\to$rel & pnt$\to$skt & rel$\to$clp & rel$\to$pnt & rel$\to$skt & skt$\to$clp & skt$\to$pnt & skt$\to$rel & Avg. \\ \midrule
    CLIP*~\cite{clip} & \multirow{4}{*}{ResNet50} & 67.9 & 84.8 & 62.9 & 69.1 & 84.8 & 62.9 & 69.2 & 67.9 & 62.9 & 69.1 & 67.9 & 84.8 & 71.2 \\
    DAPrompt*~\cite{dapl} &  & 72.4 & 87.6 & 65.9 & 72.7 & 87.6 & 65.6 & 73.2 & 72.4 & 66.2 & 73.8 & 72.9 & 87.8 & 74.8 \\
    ADCLIP*~\cite{adclip} &  & 71.7 & 88.1 & 66.0 & 73.2 & 86.9 & 65.2 & 73.6 & 73.0 & 68.4 & 72.3 & 74.2 & \textbf{89.3} & 75.2 \\
    \rowcolor[HTML]{d1cfcf}
    \textbf{UniMoS* (ours)} &  & \textbf{76.0} & \textbf{88.9} & \textbf{72.1} & \textbf{75.5} & \textbf{89.2} & \textbf{71.1} & \textbf{75.1} & \textbf{75.9} & \textbf{70.5} & \textbf{76.4} & \textbf{76.3} & 88.9 & \textbf{78.0} \\ \midrule
    CLIP*~\cite{clip} & \multirow{4}{*}{ViT-B} & 80.3 & 90.5 & 77.8 & 82.7 & 90.5 & 77.8 & 82.7 & 80.3 & 77.8 & 82.7 & 80.3 & 90.5 & 82.8 \\
    DAPrompt*~\cite{dapl} &  & 83.3 & 92.4 & 81.1 & 86.4 & 92.1 & 81.0 & 86.7 & 83.3 & 80.8 & 86.8 & 83.5 & 91.9 & 85.8 \\
    ADCLIP*~\cite{adclip} &  & 84.3 & \textbf{93.7} & 82.4 & \textbf{87.5} & \textbf{93.5} & 82.4 & \textbf{87.3} & 84.5 & 81.6 & \textbf{87.9} & 84.8 & 93.0 & 86.9 \\
    \rowcolor[HTML]{d1cfcf}
    \textbf{UniMoS* (ours)} &  & \textbf{86.2} & 93.2 & \textbf{83.2} & 86.9 & 93.2 & \textbf{83.2} & 86.8 & \textbf{86.0} & \textbf{82.8} & 87.0 & \textbf{86.2} & \textbf{93.3} & \textbf{87.3} \\ \bottomrule 
    \end{tabular}}
\end{table*}

%% file: tab/sup_tab_2.tex
% \usepackage{booktabs}
% \usepackage{multirow}
\begin{table*}[t]
    \caption{Results on Office-Home. Best results are marked in bold font. Methods with `*' are based on CLIP.}
    \label{officehome_sup}
    \resizebox{\linewidth}{!}{
    \begin{tabular}{c|c|cccccccccccc|c}
    \toprule
    Method & Backbone & A$\to$C & A$\to$P & A$\to$R & C$\to$A & C$\to$P & C$\to$R & P$\to$A & P$\to$C & P$\to$R & R$\to$A & R$\to$C & R$\to$P & Avg. \\ \midrule
    SourceOnly~\cite{resnet} & \multirow{15}{*}{ResNet50} & 34.9 & 50.0 & 58.0 & 37.4 & 41.9 & 46.2 & 38.5 & 31.2 & 60.4 & 53.9 & 41.2 & 59.9 & 46.1 \\
    ParetoDA~\cite{pareto} &  & 56.8 & 75.9 & 80.5 & 64.4 & 73.5 & 73.7 & 65.6 & 55.2 & 81.3 & 75.2 & 61.1 & 83.9 & 70.6 \\
    CLIP*~\cite{clip} &  & 51.7 & 81.5 & 82.3 & 71.7 & 81.5 & 82.3 & 71.7 & 51.7 & 82.3 & 71.7 & 51.7 & 81.5 & 71.8 \\
    SDAT~\cite{sdat} &  & 58.2 & 77.1 & 82.2 & 66.3 & 77.6 & 76.8 & 63.3 & 57.0 & 82.2 & 74.9 & 64.7 & 86.0 & 72.2 \\
    MSGD~\cite{msgd} &  & 58.7 & 76.9 & 78.9 & 70.1 & 76.2 & 76.6 & 69.0 & 57.2 & 82.3 & 74.9 & 62.7 & 84.5 & 72.3 \\
    Fixbi~\cite{fixbi} &  & 58.1 & 77.3 & 80.4 & 67.7 & 79.5 & 78.1 & 65.8 & 57.9 & 81.7 & 76.4 & 62.9 & 86.7 & 72.7 \\
    CST~\cite{cst} &  & 59.0 & 79.6 & 83.4 & 68.4 & 77.1 & 76.7 & 68.9 & 56.4 & 83.0 & 75.3 & 62.2 & 85.1 & 72.9 \\
    ATDOC~\cite{atdoc} &  & 60.2 & 77.8 & 82.2 & 68.5 & 78.6 & 77.9 & 68.4 & 58.4 & 83.1 & 74.8 & 61.5 & 87.2 & 73.2 \\
    KUDA~\cite{kuda} &  & 58.2 & 80.0 & 82.9 & 71.1 & 80.3 & 80.7 & 71.3 & 56.8 & 83.2 & 75.5 & 60.3 & 86.6 & 73.9 \\
    DAPrompt*~\cite{dapl} &  & 54.1 & 84.3 & 84.8 & 74.4 & 83.7 & 85.0 & 74.5 & 54.6 & 84.8 & 75.2 & 54.7 & 83.8 & 74.5 \\
    EIDCo~\cite{eidco} &  & \textbf{63.8} & 80.8 & 82.6 & 71.5 & 80.1 & 80.9 & 72.1 & 61.3 & 84.5 & \textbf{78.6} & 65.8 & 87.1 & 75.8 \\
    ICON~\cite{icon} &  & 63.3 & 81.3 & 84.5 & 70.3 & 82.1 & 81.0 & 70.3 & \textbf{61.8} & 83.7 & 75.6 & \textbf{68.6} & 87.3 & 75.8 \\ 
    ADCLIP*~\cite{adclip} &  & 55.4 & 85.2 & 85.6 & 76.1 & 85.8 & 86.2 & \textbf{76.7} & 56.1 & 85.4 & 76.8 & 56.1 & 85.5 & 75.9 \\
    PADCLIP*~\cite{padclip} &  & 57.5 & 84.0 & 83.8 & \textbf{77.8} & 85.5 & 84.7 & 76.3 & 59.2 & 85.4 & 78.1 & 60.2 & 86.7 & 76.6 \\ 
    \rowcolor[HTML]{d1cfcf} 
    \textbf{UniMoS* (ours)} &  & 59.5 & \textbf{89.4} & \textbf{86.9} & 75.2 & \textbf{89.6} & \textbf{86.8} & 75.4 & 58.4 & \textbf{87.2} & 76.9 & 59.5 & \textbf{89.7} & \textbf{77.9} \\ \bottomrule
    SourceOnly~\cite{vit} & \multirow{8}{*}{ViT-B} & 54.7 & 83.0 & 87.2 & 77.3 & 83.4 & 85.5 & 74.4 & 50.9 & 87.2 & 79.6 & 53.8 & 88.8 & 75.5 \\
    CLIP*~\cite{clip} &  & 67.8 & 89.0 & 89.8 & 82.9 & 89.0 & 89.8 & 82.9 & 67.8 & 89.8 & 82.9 & 67.8 & 89.0 & 82.4 \\
    TVT~\cite{tvt} &  & 74.9 & 86.8 & 89.5 & 82.8 & 88.0 & 88.3 & 79.8 & 71.9 & 90.1 & 85.5 & 74.6 & 90.6 & 83.6 \\
    DAPrompt*~\cite{dapl} &  & 70.6 & 90.2 & 91.0 & 84.9 & 89.2 & 90.9 & 84.8 & 70.5 & 90.6 & 84.8 & 70.1 & 90.8 & 84.0 \\
    SSRT~\cite{ssrt} &  & 75.2 & 89.0 & 91.1 & 85.1 & 88.3 & 89.9 & 85.0 & 74.2 & 91.2 & 85.7 & 78.6 & 91.8 & 85.4 \\
    ADCLIP*~\cite{adclip} &  & 70.9 & 92.5 & 92.1 & 85.4 & 92.4 & \textbf{92.5} & \textbf{86.7} & 74.3 & \textbf{93.0} & \textbf{86.9} & 72.6 & 93.8 & 86.1 \\
    PADCLIP*~\cite{padclip} &  & \textbf{76.4} & 90.6 & 90.8 & \textbf{86.7} & 92.3 & 92.0 & 86.0 & \textbf{74.5} & 91.5 & \textbf{86.9} & \textbf{79.1} & 93.1 & 86.7 \\
    \rowcolor[HTML]{d1cfcf}
    \textbf{UniMoS* (ours)} &  & 74.9 & \textbf{94.0} & \textbf{92.5} & 86.4 & \textbf{94.3} & \textbf{92.5} & 86.0 & 73.9 & \textbf{93.0} & 86.4 & 74.2 & \textbf{94.5} & \textbf{86.9} \\ \midrule
    CLIP*~\cite{clip} & \multirow{4}{*}{ViT-L} & 74.2 & 93.1 & 93.3 & 87.3 & 93.1 & 93.3 & 87.3 & 74.2 & 93.3 & 87.3 & 74.2 & 93.1 & 87.0 \\
    DAPrompt*~\cite{dapl} &  & 77.3 & 94.6 & 94.3 & 88.6 & 94.6 & 94.0 & 88.8 & 76.8 & 94.0 & 89.0 & 77.8 & 94.4 & 88.7 \\
    ADCLIP*~\cite{adclip} &  & 80.3 & 95.4 & \textbf{95.7} & \textbf{90.9} & 95.5 & \textbf{95.2} & \textbf{90.1} & 79.6 & 95.1 & \textbf{90.8} & 81.1 & 95.9 & 90.5 \\
    \rowcolor[HTML]{d1cfcf}
    \textbf{UniMoS* (ours)} &  & \textbf{80.9} & \textbf{96.2} & 95.1 & 90.1 & \textbf{96.1} & 95.1 & 90.0 & \textbf{81.4} & \textbf{95.2} & 89.9 & \textbf{81.6} & \textbf{96.3} & \textbf{90.7} \\ \bottomrule
    \end{tabular}}
\end{table*}

%% file: tab/sup_tab_3.tex
% \usepackage{booktabs}
% \usepackage{multirow}
\begin{table*}[t]
    \caption{Computation analysis. Best results are marked in bold font. Methods with `*' are based on CLIP.}
    \centering
    \label{computation_sup}
    \resizebox{0.9\linewidth}{!}{
    \begin{tabular}{c|c|c|ccccc|c}
    \toprule
    Method & Dataset & Backbone & Param. & Throughput (imges/s) & Train time & FLOPs & GPU mem. & Acc. \\ \midrule
    DAPrompt* &  &  & 1.2M & 244 & 4.3H & 11.3G & 6.9G & 86.9 \\
    FixBi &  &  & 86.1M & 102 & 5.5H & 15.73G & 17.0G & 87.2 \\
    CAN &  &  & 42.5M & 31 & 10.5H & 7.9G & 11.3G & 87.2 \\ 
    PADCLIP* &  &  & - & - & 23.5H & - & - & \textbf{88.5} \\
    \rowcolor[HTML]{d1cfcf}
    \textbf{UniMoS* (ours)} & \multirow{-5}{*}{VisDA} & \multirow{-5}{*}{ResNet101} & \textbf{0.79M} & \textbf{2667} & \textbf{0.5H} & \textbf{\textless{}0.01G} & \textbf{1.8G} & 88.1 \\ \midrule
    PMTrans & \multirow{3}{*}{DomainNet} & Swin & 89.5M & 46 & 30H & 15.2G & 19.3G & 52.4 \\ \cmidrule{3-3}
    DAPrompt* &  &  & 34.5M & 31 & 7.9H & 73.9G & 22.5G & 59.8 \\ 
    \rowcolor[HTML]{d1cfcf}
    \textbf{UniMoS* (ours)} &  & \multirow{-2}{*}{ViT-B} & \textbf{0.79M} & \textbf{2827} & \textbf{0.4H} & \textbf{\textless{}0.01G} & \textbf{2.9G} & \textbf{63.6} \\ \bottomrule
    \end{tabular}}
\end{table*}